\documentclass[10pt,twocolumn,letterpaper]{article}

\usepackage{iccv}
\usepackage{times}
\usepackage{epsfig}
\usepackage{graphicx}
\usepackage{amsmath}
\usepackage{amssymb}
\usepackage[normalem]{ulem}
\usepackage{nicefrac}
\usepackage{booktabs}
\usepackage{placeins}
\usepackage{enumitem}
\usepackage[accsupp]{axessibility}  

\usepackage[pagebackref=true,breaklinks=true,letterpaper=true,colorlinks,bookmarks=false]{hyperref}

\iccvfinalcopy %

\def\iccvPaperID{10364} %

\newcommand{\rn}{\text{ResNet}}
\newcommand{\ione}{\text{ILSVRC-2012}}
\newcommand{\itwentyone}{\text{ImageNet-21k}}
\newcommand{\jft}{\text{JFT-300M}}
\newcommand{\ir}{\text{ImageNet-R}}
\newcommand{\ic}{\text{ImageNet-C}}
\newcommand{\ia}{\text{ImageNet-A}}
\newcommand{\cs}{\text{Conflict Stimuli}}

\begin{document}

\title{Understanding Robustness of Transformers for Image Classification}

\newcommand{\ft}{$^*$}

\author{Srinadh Bhojanapalli\ft, Ayan Chakrabarti\ft, Daniel Glasner\ft, Daliang Li\ft, Thomas Unterthiner\ft, Andreas Veit\ft\\
Google Research\\
\tt\small \{bsrinadh, ayanchakrab, dglasner, daliangli, unterthiner, aveit\}@google.com}

\maketitle
\ificcvfinal\thispagestyle{empty}\fi

\renewcommand{\thefootnote}{*}
\footnotetext{All authors contributed equally.}

\begin{abstract}
Deep Convolutional Neural Networks (CNNs) have long been the architecture of choice for computer vision tasks. Recently, Transformer-based architectures like Vision Transformer (ViT) have matched or even surpassed ResNets for image classification. However, details of the Transformer architecture --such as the use of non-overlapping patches-- lead one to wonder whether these networks are as robust. In this paper, we perform an extensive study of a variety of different measures of robustness of ViT models and compare the findings to \rn\ baselines. We investigate robustness to input perturbations as well as robustness to model perturbations. We find that when pre-trained with a sufficient amount of data, ViT models are at least as robust as the \rn \ counterparts on a broad range of perturbations. We also find that Transformers are robust to the removal of almost any single layer, and that while activations from later layers are highly correlated with each other, they nevertheless play an important role in classification.

\end{abstract}

\section{Introduction}\label{sec:intro}

Convolutions have served as the building blocks of computer vision algorithms in nearly every application domain---with their property of spatial locality and translation invariance mapping naturally to the characteristics of visual information. %
Neural networks for vision tasks adopted the use of convolutional layers quite early on \cite{fukushima1982neocognitron, lecun1998gradient}, and since their resurgence with Krizhevsky et al.'s work~\cite{krizhevsky2012imagenet}, all modern networks for vision have been convolutional~\cite{simonyan2014very, szegedy2015going, he2016deep, howard2017mobilenets, huang2017densely, hu2018squeeze, tan2019efficientnet}---with innovations such as residual~\cite{he2016deep} connections being applied to a backbone of convolutional layers. Given their extensive use, convolutional networks have been the subject of significant analysis---both empirical~\cite{szegedy2013intriguing} and analytical~\cite{girshick2015deformable, battaglia2018relational}.

Recently, after seeing tremendous success in language tasks~\cite{vaswani2017attention,devlin2019bert,brown2020language}, researchers have been exploring a variety of avenues for deploying attention-based \emph{Transformer} networks~\cite{chen2020generative, dosovitskiy2020image, touvron2020deit, khan2021transformers}---and other attention-based architectures~\cite{Xu2015show,Wang2018nonlocal,Ramachandran2019sasa,Locatello2020slotattention,Zhao_2020_CVPR}---in computer vision. Transformers are also gaining popularity in vision and language tasks~\cite{sun2019videobert, lu2019vilbert, tan2019lxmert, chen2019uniter, lu202012, radford2021learning}.

In this paper, we focus on one particular Transformer architecture---the Visual Transformer (ViT) introduced by Dosovitskiy et al.~\cite{dosovitskiy2020image}---because it was shown to achieve better performance than state-of-the-art residual networks (ResNets)~\cite{he2016deep} of similar capacity, when both are pre-trained on sufficiently large datasets~\cite{kolesnikov2019big}, such as \jft~\cite{sun2017revisiting}. We also focus on ViT because, unlike other Transformer models for vision, their architecture consists solely of Transformer layers.

Dosovitskiy et al.'s results~\cite{dosovitskiy2020image} tell us that such an architecture is preferable in terms of performance, given enough training data to overcome the lack of an inductive bias through convolutions. But, the pure attention-based mechanism by which ViT models process their inputs is a significant departure from the familiar decade-long ubiquitous use of convolution networks for computer vision. In this paper, we seek to gain a better understanding of how these architectures behave---in terms of their robustness against perturbations to inputs and to the model parameters themselves---and build up an understanding of these models that parallels our knowledge about convolution.

\begin{figure}[!t]
\centering
\includegraphics[width=\columnwidth]{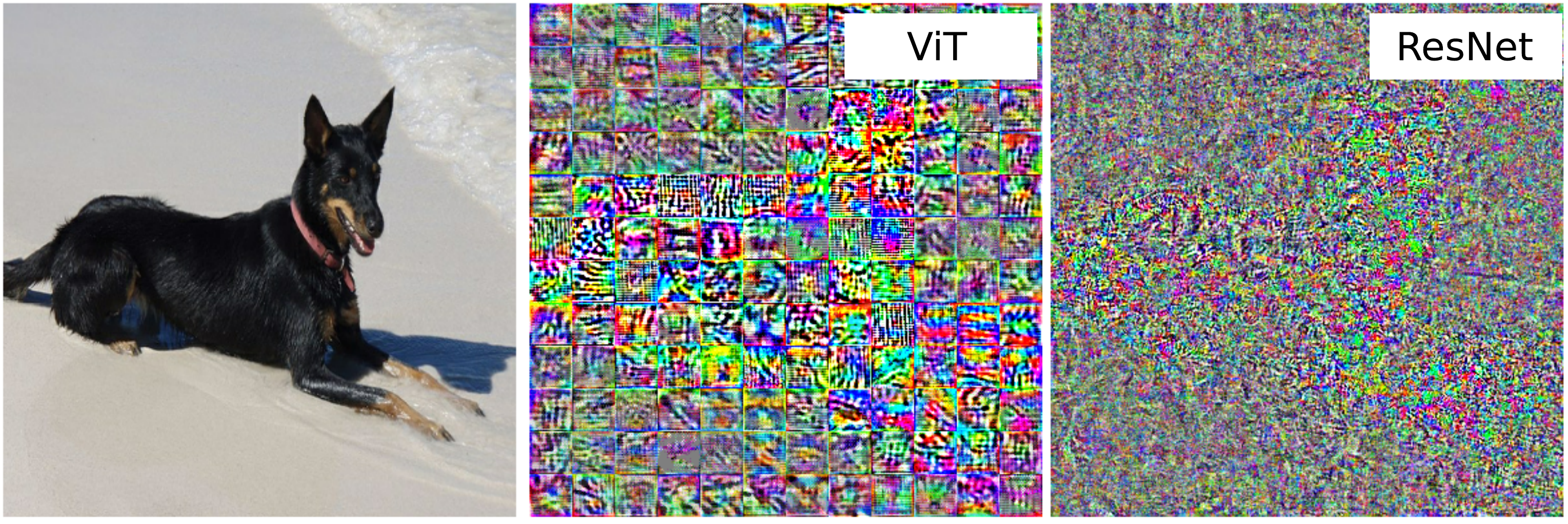}
\caption{\textbf{Transformers vs.\ ResNets}. While they achieve similar performance for image classification, Transformer and ResNet architectures process their inputs very differently. Shown here are adversarial perturbations computed for a Transformer and a ResNet model, which are qualitatively quite different.\label{fig:teaser}}
\end{figure}

We begin with an exhaustive set of experiments comparing the performance of various ViT model variants, under different perturbations to their image inputs, to similarly sized and trained ResNet architectures~\cite{he2016deep, kolesnikov2019big}. Perturbations range from natural variations~\cite{hendrycks2019natural, hendrycks2019robustness, hendrycks2020many} to adversarial perturbations~\cite{szegedy2013intriguing,goodfellow2014explaining,hendrycks2019natural,hsieh2019robustness} and spatial transformations~\cite{engstrom2019exploring}. We also evaluate texture and shape bias~\cite{geirhos2018imagenet}.

We then turn our attention to the action of the ViT models themselves, analyzing the evolution of information through the cascade of Transformer layers and the redundancies in their internal representations via correlation analysis and lesion studies---as has been done in the past for  Transformers for language tasks~\cite{voita2019analyzing,press2020improving,mandava2020pay,dong2021attention} and for ResNets for vision tasks~\cite{veit2016residual, greff2017highway}. Moreover, since it is known that self-attention can learn to mimic convolutions~\cite{Cordonnier2020on}, we also investigate the effect of enforcing spatial locality in the attention mechanism of the Transformer layers of ViT models.

Our investigations provide researchers and practitioners with a deeper understanding of how this new class of deep network architectures work, the range of applications to which they may be deployed, and provide potential avenues for how they may be improved in terms of performance or efficiency. Our contributions are as follows:
\begin{itemize}[leftmargin=1em,itemsep=0pt,parsep=0pt,topsep=0pt,partopsep=0pt]
    \item We measure robustness of ViT models of different sizes that are pre-trained on different datasets and compare them to corresponding \rn \ baselines.
    \item We measure robustness with respect to input perturbations, and find that ViTs pre-trained on sufficiently large datasets tend to be generally at least as robust as their \rn \ counterparts.
    \item We measure robustness with respect to model perturbations, and find that ViTs are robust to the removal of almost any single layer, and that later layers provide only limited updates to the representations of individual patches, but focus on consolidating information in the CLS token.
\end{itemize}

\begin{table*}[!t]
\centering
\begin{tabular}{lcccccccc}\toprule
\textbf{Model} & ViT-B & ViT-L & ViT-H & ResNet-50x1 & ResNet-101x1 & ResNet-50x3 & ResNet-101x3 & ResNet-152x4 \\\midrule
\textbf{\# Params} & 86M & 307M & 632M & 23M & 45M & 207M & 401M & 965M\\\bottomrule
\end{tabular}\vspace{0.25em}

\caption{\textbf{Architectures.} Model architectures used in our experiments along with the number of learnable parameters for each. }
\label{table:model_variants}
\end{table*}

\begin{figure*}[!th]
\includegraphics[width=\textwidth]{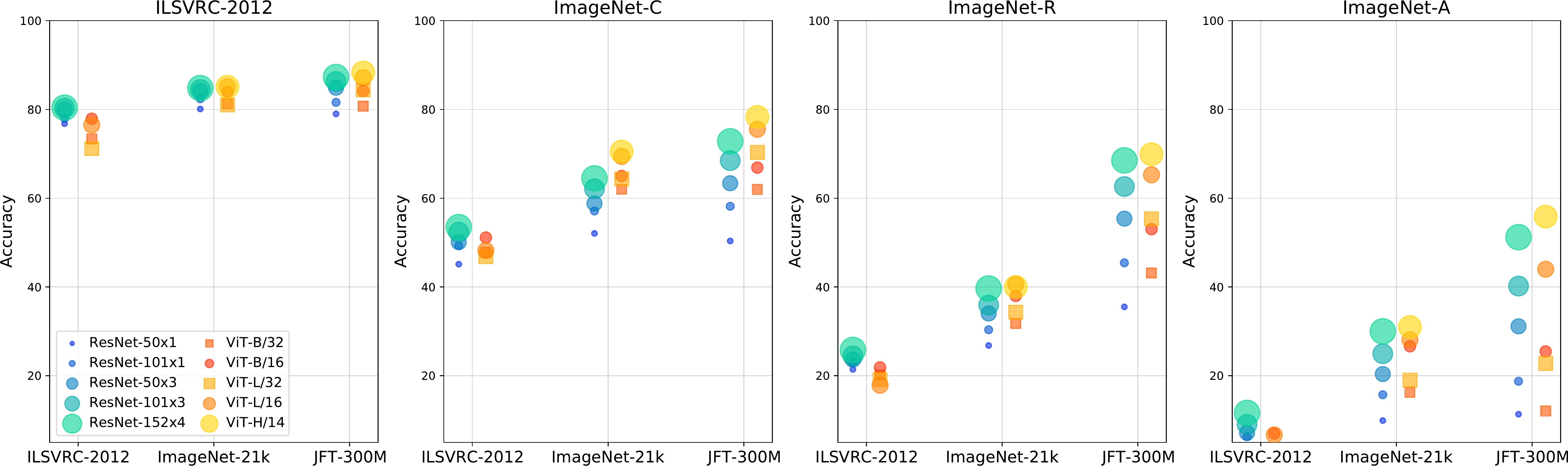}
\caption{\textbf{Robustness Benchmarks}. Accuracy of ViT and \rn\ models on \ione\ (clean), \ic, \ir\ and \ia{}. For \ic\ the accuracy is averaged across all corruption types and severity levels. We observe that (i) relative accuracy on \ione{} is generally predictive of relative accuracy on the perturbed datasets, and that when trained on sufficient data, the accuracy of ViT models (ii) outperforms \rn{}s, and (iii) scales better with model size. Marker size related to model size. Detailed results for \ic\ can be found in Appendix~\ref{appendix:detailed_results}.}\label{fig:imagenet-c-r-a-cs}
\end{figure*}

\section{Preliminaries}\label{sec:related_work}
\subsection{Transformers}
Self-attention based Transformer architectures were introduced in \cite{vaswani2017attention}, where they showed superior performance on machine translation. They have since been applied successfully to many tasks in NLP. Notably \cite{devlin2019bert,Brown2020gpt3} have shown that combined with pre-training, these models achieve nearly human performance on a wide range of NLP tasks.

The input to Transformer models is a sequence of vectors---typically embeddings of input tokens---that are processed by a stack of  Transformer blocks. Each block consists of 1) a multi-head self-attention layer that aggregates information across tokens using dot-product attention; and 2) a tokenwise feed-forward (MLP) layer.  %
 Both use layer normalization and residual connections.

\paragraph{Vision Transformer}
ViT \cite{dosovitskiy2020image} uses the same Transformer architecture discussed above. They key difference comes in the image pre-processing layer. This layer partitions the image into a sequence of non-overlapping patches followed by a learned linear projection. For example, a $384 \times 384$ image can be broken into $16 \times 16$ patches resulting in a sequence length of $16^2$. This is accomplished using a 2D convolution, where the number of filters determines the hidden size of the sequence input to the Transformer. Following \cite{devlin2019bert} ViT also appends a special CLS token to the input, whose representation is used for final classification.

\subsection{Model Variants}\label{sec:model_variants}
In order to better understand and contrast ViTs and \rn{}s, we evaluate a range of models from each architecture family. We follow~\cite{dosovitskiy2020image} and use models which vary in the number of parameters, their input patch-size, and in the datasets on which they were pre-trained. Table~\ref{table:model_variants} summarizes the sizes of the models used in our experiments. We append ``/$x$'' to model-names to denote models that take patches of size $x\cdot x$ as input, and use model variants that were pre-trained either on \ione, with $\sim$~1.3 million images, on \itwentyone, with $\sim$~12.8 million images, or on \jft~\cite{sun2017revisiting} which contains around 375M labels for 300M images. All models are finetuned on \ione.  We obtained saved parameter checkpoints for the ViT models from the authors of~\cite{dosovitskiy2020image}, and those for the \rn\ models from the authors of~\cite{kolesnikov2019big}.

\section{Robustness to Input Perturbations}\label{sec:robustness_to_input_perturbations}

In this section we compare the robustness to input perturbations of ViT models to \rn{}s. We do this by measuring performance of a range of representatives from each architecture family, as described in Sec.~\ref{sec:model_variants}.
To capture different aspects of robustness we rely on different, specialized benchmarks \ic, \ir\ and \ia. We also pit our models against different types of adversarial attacks. Finally, we explore the texture bias of ViTs.

\subsection{Natural Corruptions}\label{sec:imagenet-c}
So called ``natural" or ``common" perturbation benchmarks provide an important yardstick for estimating real-world performance in the presence of naturally occurring image corruptions~\cite{hendrycks2019robustness, gulshad2020adversarial, laugros2019adversarial}. Robustness to such perturbations can be important for example in safety-critical applications. We use \ic, a benchmark introduced in~\cite{hendrycks2019robustness} to evaluate ViT's robustness to natural corruptions. \ic\ includes 15 types of algorithmically generated corruptions, grouped into 4 categories: `noise', `blur', `weather', and `digital'. Each corruption type has five levels of severity, resulting in 75 distinct corruptions. 

Our results, averaged over all corruptions and all severities, are shown in the second column of Fig.~\ref{fig:imagenet-c-r-a-cs}. More granular results can be found in Appendix~\ref{appendix:detailed_results}. We find that the size of the pre-training dataset has a fundamental effect on the robustness of ViTs. When the training set is small, the ViTs are less robust compared to \rn{}s of comparable sizes, and increasing the size of the ViTs does not lead to better robustness. This is consistent with performance on the clean set, and with the observations of~\cite{dosovitskiy2020image} about the inductive bias of convolutions being useful when pre-training data is limited. However, when the training data is \itwentyone, we observe stronger robustness for most ViT models. This effect becomes even more pronounced when the models are pre-trained on \jft, where ViTs show better robustness against most corruptions compared to \rn{}s. Moreover, in the larger pre-training data regime, performance gains can be achieved for ViT models by increasing the model size or by decreasing the patch size (and thus increasing the amount of computation).

\subsection{Real-World Distribution Shifts}\label{sec:imagenet-r}
Robustness to distribution shift, can be measured in different ways. Here, we evaluate ViT models on \ir~\cite{hendrycks2020many}, a dataset with different ``renditions" of \ione \ classes. An advantage of \ir\ is that the renditions are real-world, naturally occurring variations, such as paintings or embroidery, with textures and local image statistics which differ from those of ImageNet images. 

Despite the fundamental difference in the nature of the perturbations in \ir\ and \ic, the models' behavior on \ir\ is similar, as shown in Fig.~\ref{fig:imagenet-c-r-a-cs}. Again, ViTs underperform \rn{}s when the pre-training data is small and starts to outperform them when pre-trained on larger datasets. The benefit of larger model sizes is also more clear on larger datasets, especially for ViTs.  

The behavior of our baseline \rn\ models is in line with those observed in Appendix G of~\cite{kolesnikov2019big}, where they are evaluated on objects out-of-context. The authors of~\cite{kolesnikov2019big} create a dataset of
foreground objects corresponding to \ione\ classes pasted onto miscellaneous backgrounds. They find that when using more pre-training data, better performance of the larger model on \ione\ translates to better out-of-context performance. 

Our finding that more pre-training data improves performance on out-of-distribution data is also in line with the findings in NLP. Hendrycks et al.~\cite{hendrycks2020pretrained} show that pre-trained Transformers improve robustness on a variety of out-of-distribution NLP benchmarks. One of their interesting findings is that for NLP, larger models are not always better. We observe a similar phenomenon for ViTs pre-trained on \ione, but not for ViTs pre-trained on \itwentyone\ or on \jft.

\begin{figure*}[!t]
\includegraphics[width=\textwidth]{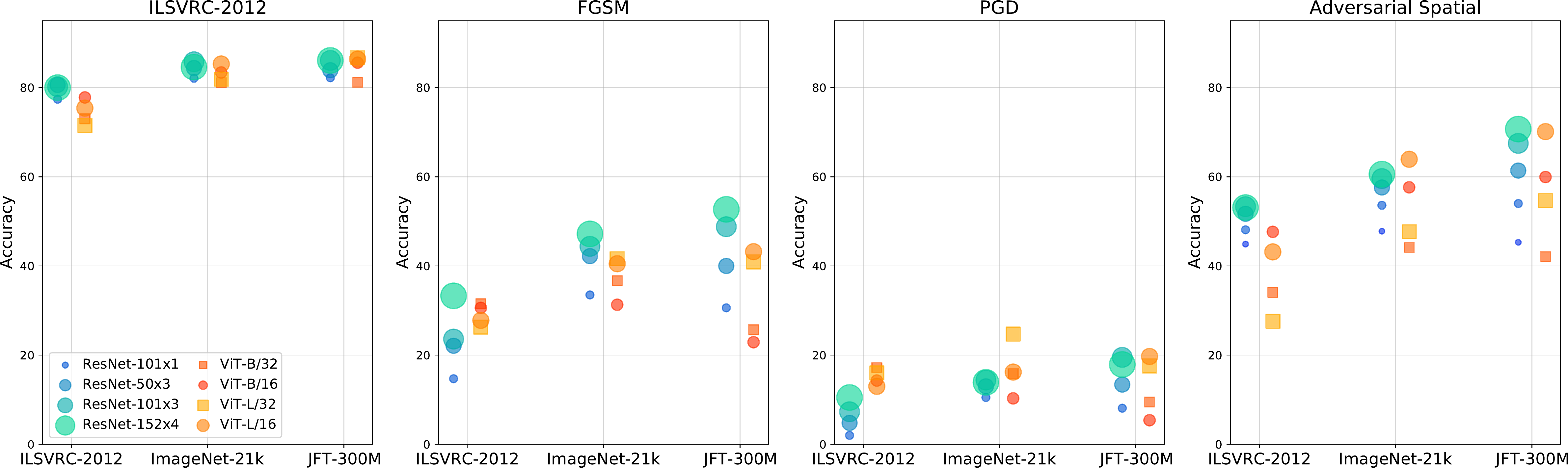}
\caption{\textbf{Adversarial Perturbations}. Accuracy on a subset of 1000 images in \ione\ validation of ViT and \rn\ models, on clean images \emph{(left)} vs.\ those subject to model-specific adversarial attacks: FGSM and PGD-based perturbations \emph{(middle)}, and spatial (rotation and translation) transformations \emph{(right)}. (We omit ViT-H/14 here, since it expects a different input image resolution than the other models.) \rn\ models are more robust to the simpler FGSM attack than their ViT counterparts, but this advantage disappears for the more successful PGD attacks. For spatial attacks, the $16\times16$ ViT models exhibit equivalent robustness to \rn{}s of comparable size, but ViT models with the larger patch-size of $32\times32$ fare worse.}\label{fig:adversarial}
\end{figure*}

\subsection{Natural Adversarial Examples}\label{sec:imagenet-a}
Adversarial robustness is usually measured by considering the worst-case perturbation within a small radius in image space. We explore ViTs performance on such perturbations in Sec.~\ref{sec:robust-advers-pert}. In contrast, the so called ``natural adversarial" examples of Hendrycks et al.~\cite{hendrycks2019natural} are unmodified real-world images which were found by filtering with a trained ResNet-50 model, and have been shown to transfer to other models. In contrast to \ic\ and \ir, the local statistics of these images is similar to ImageNet images.  

Our results on \ia\ are shown in the right column of Fig.~\ref{fig:imagenet-c-r-a-cs}. We find that ViTs, despite having a dramatically different architecture compared to ResNet-50, are susceptible to the same natural adversarial images. Again we find that larger pre-training datasets are beneficial to ViT models, which start to outperform \rn{}s when both are pre-trained on \jft. This finding should be taken with a grain of salt, since the adversarial selection process is based on a ResNet-50, so the examples might be harder for \rn{}s by design. 

\subsection{Robustness and Model Size}\label{sec:scaling_discussion}
\begin{figure}[!t]
\hspace{0.7cm}\includegraphics[width=0.8\columnwidth]{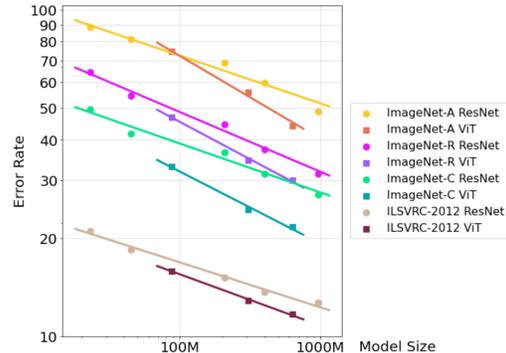}
\caption{\textbf{Scaling}. Performance of ViT and \rn\ models as a function of the number of model parameters.  All models are pre-trained on \jft \ and fine-tuned on \ione. We see consistent trends across different input perturbations: scaling up ViTs provides better robustness gains than scaling up \rn{}s.}\label{fig:scaling}
\end{figure}

On sufficiently large datasets, it is well known that for a fixed architecture, larger models lead to better quality. Kaplan et al.~\cite{kaplan2020scaling} demonstrated that such improvements on Transformers trained on large NLP datasets follow clear and predictable power laws. In previous subsections, we found that in addition to clean performance, the robustness of ViTs and \rn{}s against various input perturbations also improves with model size. The gap between large and small models grows as the dataset becomes bigger. It is therefore interesting to evaluate the relation between a models' robustness and its size, when pre-trained on the largest dataset, \jft. The results are summarized in Fig.~\ref{fig:scaling}. 

We find that the error rates follow consistent trends when scaling up the model size, across up to two orders of magnitude. This holds true on different robustness benchmarks, as well as the clean \ione\ validation set. We also note that ViTs exhibit more favorable scaling compared to \rn. This suggests that given a sufficiently large pre-training dataset, such as \jft, the gap in robustness between ViTs and \rn{}s will further increase as the models become bigger and bigger. Note that this advantage of ViTs is only realized when the pre-training dataset is sufficiently large. In Appendix~\ref{appendix:i21k_scaling} we show that when pre-trained on \itwentyone, ViTs' robustness does not scale better than \rn{}s'.

We also find that for the same model family, the slope in the error rate vs. model size relation remains relatively consistent across different datasets, despite their drastically different characteristics. This suggests that the scaling trends we discovered might generalize to a broader set of evaluation datasets and tasks. 

\begin{figure*}[!t]
\includegraphics[width=\textwidth]{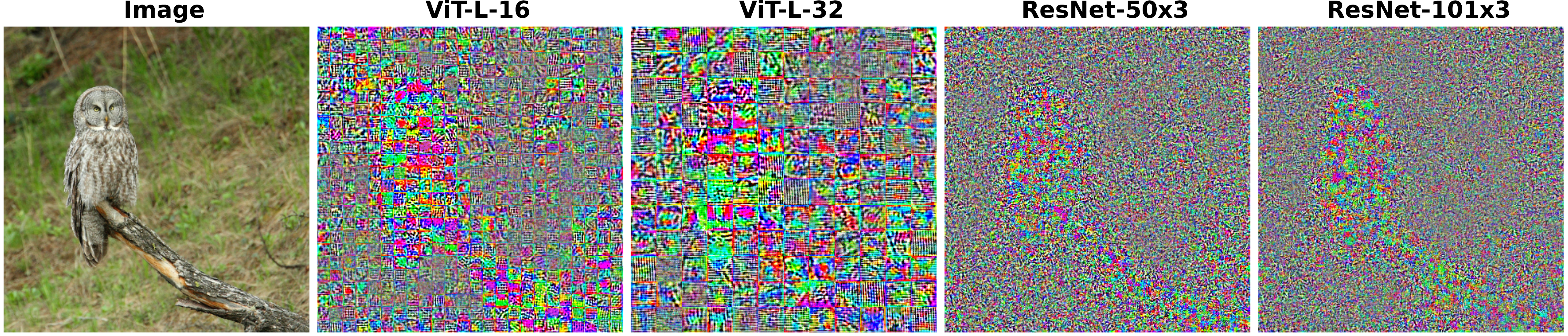}
\caption{\textbf{Example Perturbations}. For example images from the ILSVRC 2012 validation set, we illustrate the perturbations computed using PGD for two ViT models and two ResNet models (we use models pre-trained on JFT-300M). The perturbations are visualized as images by linearly transforming their intensity from the original range of $[-1, 1]$ to $[0, 255]$.}\label{fig:pgdimg}
\end{figure*}

\subsection{Adversarial Perturbations}\label{sec:robust-advers-pert}
Most deep neural network models are vulnerable to \emph{adversarial perturbations}~\cite{szegedy2013intriguing}---extremely small, but carefully crafted, perturbations to the input, that cause a model to produce incorrect predictions. In NLP, Hsieh et al.~\cite{hsieh2019robustness} have shown that attention-based models tend to be more robust to such perturbations than other architectures (such as recurrent networks). In this section, we evaluate the robustness of various ViT and ResNet models for image classification to adversarial perturbations.

We consider perturbations with an $L_{\infty}$ norm of one gray level, computed with knowledge of the model architecture and weights (i.e., white-box attacks). We use two standard approaches to compute these perturbations: the Fast Gradient Sign Method (FGSM)~\cite{goodfellow2014explaining} and Projected Gradient Descent (PGD)~\cite{pgd}, using eight iterations with a step size of $1/8$ gray levels for the latter. Figure~\ref{fig:adversarial} reports the accuracies over a subset of 1000 images from the \ione\  validation set, on the original images and after adding perturbations computed using both methods.

We see that the performance of all models degrades with these perturbations, and as expected, PGD is more successful than FGSM. Also, we find that larger models tend to be more robust than smaller ones, and that pre-training on larger datasets improves robustness to adversarial perturbations. 
Interestingly, among models that are trained only on \ione, the Transformer models appear to be more robust than ResNet models of equivalent size---quite a bit more so with perturbations computed using PGD. Among models trained with a medium amount of training data (pre-trained on \itwentyone), we find that ResNet models are more robust to the simpler FGSM attack than their Transformer counterparts, but the opposite is true for PGD attacks. Finally, among models trained with the most training data, robustness to FGSM appears largely to be monotonic with model size. PGD attacks are again more successful, but here, there appear to be diminishing returns with model size once one crosses 300 million parameters.

An interesting observation is that the \emph{relative} robustness of ViT models to their ResNet counterparts appears to be lower for attacks with FGSM than PGD. This is likely due to the presence of the single large linear patch-embedding layer at the beginning of all the ViT models, which causes the single-iteration gradients used by FGSM to better correspond to a pattern coordinated across larger spatial regions. This disadvantage disappears with multiple PGD iterations.

We visualize example patterns computed (using PGD) for Transformer and ResNet models in Fig.~\ref{fig:pgdimg}, and find them to be qualitatively quite different. For all models the perturbations have the highest magnitudes around the foreground objects. For ViT, there is a clear alignment of the patterns to the patch partition boundaries. In contrast, the patterns for ResNet models
are more spatially incoherent. 

Finally, we find that adversarial patterns do not transfer over between ViT and ResNet architectures---i.e., patterns computed using ViT models rarely degrade the performance of ResNet models and vice-versa (see Table~\ref{tab:xfer} and details in Appendix~\ref{appendix:adversarial-cross-over}). 
This stands in contrast to our observations with \emph{natural} adversarial images described in Sec.~\ref{sec:imagenet-a}. 

\begin{table}[!t]
\centering
\begin{tabular}{lcc}\toprule
& \bf ViT$\rightarrow$RN & \bf RN$\rightarrow$ViT\\\midrule
ViT-B/16 vs.\ RN-101x1 & 79.7\% (-2.5) & 85.2\% (-0.4)\\
ViT-B/32 vs.\ RN-50x3  & 82.2\% (-1.7) & 80.9\% (-0.3)\\
ViT-L/16 vs.\ RN-101x3 & 84.3\% (-1.8) & 85.8\% (-0.6)\\
ViT-L/32 vs.\ RN-152x4 & 85.4\% (-0.7) & 86.5\% (-0.2)\\\bottomrule
\end{tabular}\vspace{0.25em}
\caption{\textbf{Transferability.} Accuracy when evaluating adversarial perturbations computed (with PGD) using ViT on ResNet models, and vice-versa. 
All models are pre-trained on JFT-300M. Numbers in parenthesis indicate difference from accuracy on unperturbed images. The results indicate that adversarial perturbations do not transfer well between ViT and ResNet models. Additional details can be found in Appendix~\ref{appendix:adversarial-cross-over}.} 
\label{tab:xfer}
\end{table}

\subsection{Adversarial Spatial Perturbations}\label{sec:robust-spatial-advers-pert}
We now measure the spatial robustness of these models, following the approach of Engstrom et al.~\cite{engstrom2019exploring} who explore the landscape of spatial robustness using \emph{adversarial examples}. In this setting, the adversary's attack is chosen from a given range of translations and rotations. The attack succeeds if any rotated and translated version of the image is misclassified. These attacks are chosen to particularly test the differences in input processing of these models. For example, ViTs' use of large non-overlapping patches could increase their sensitivity to subpatch-sized shifts

We test the performance of ViT and \rn\ models under grid attacks, (grid search over a discrete set of rotations and translations), as these were found to be significantly more powerful than any of the other attacks considered in~\cite{engstrom2019exploring}. We consider 9 equally spaced values each, for horizontal and vertical translations in the range $[-16, 16]$ pixels, and 31 equally spaced values for rotations in the range $[-30^{\circ}, 30^{\circ}]$. Following~\cite{engstrom2019exploring}, when rotating and translating the images, we fill the empty space with zeros (black pixels). We chose the translation ranges to span the largest patch size ($32 \times 32$) used by any of the ViT models.

We present the results averaged over 1000 images from the validation set of \ione\ in the right column of Fig.~\ref{fig:adversarial}, and find both ViT and \rn\ models to be susceptible to spatial attacks. 
Surprisingly, ViT models with a patch size of $16 \times 16$ mostly maintain their positions relative to the \rn\ models, indicating they are no more susceptible to spatial adversarial attacks. In contrast, the performance of ViT models that use a larger patch size of $32 \times 32$, degrades much more than the comparable \rn\ models. We conclude that ViT models with smaller patch size, seem to be as robust to translations and rotations as comparable \rn{}s. However, ViT models with larger patch sizes tend to be more susceptible to spatial attacks.

\subsection{Texture Bias}\label{sec:texture_and_shape}
Geirhos et al.~\cite{geirhos2018imagenet} observe that (unlike humans) ImageNet-trained CNNs tend to rely on texture more than on shape for image classification. They further report that reducing the texture bias leads to improved robustness to previously unseen image distortions.
We evaluate the texture bias of ViT models and compare it to ResNets using the \cs\ benchmark of~\cite{geirhos2018imagenet}. This dataset is generated by combining 160 images of objects with white background and 48 texture images using style transfer, resulting in 1280 test images with different (possibly conflicting) shape and texture combinations. The fraction of examples in this dataset that are classified correctly by their shape determines the shape accuracy of a model. %

\begin{figure}[!t]
\includegraphics[width=\columnwidth]{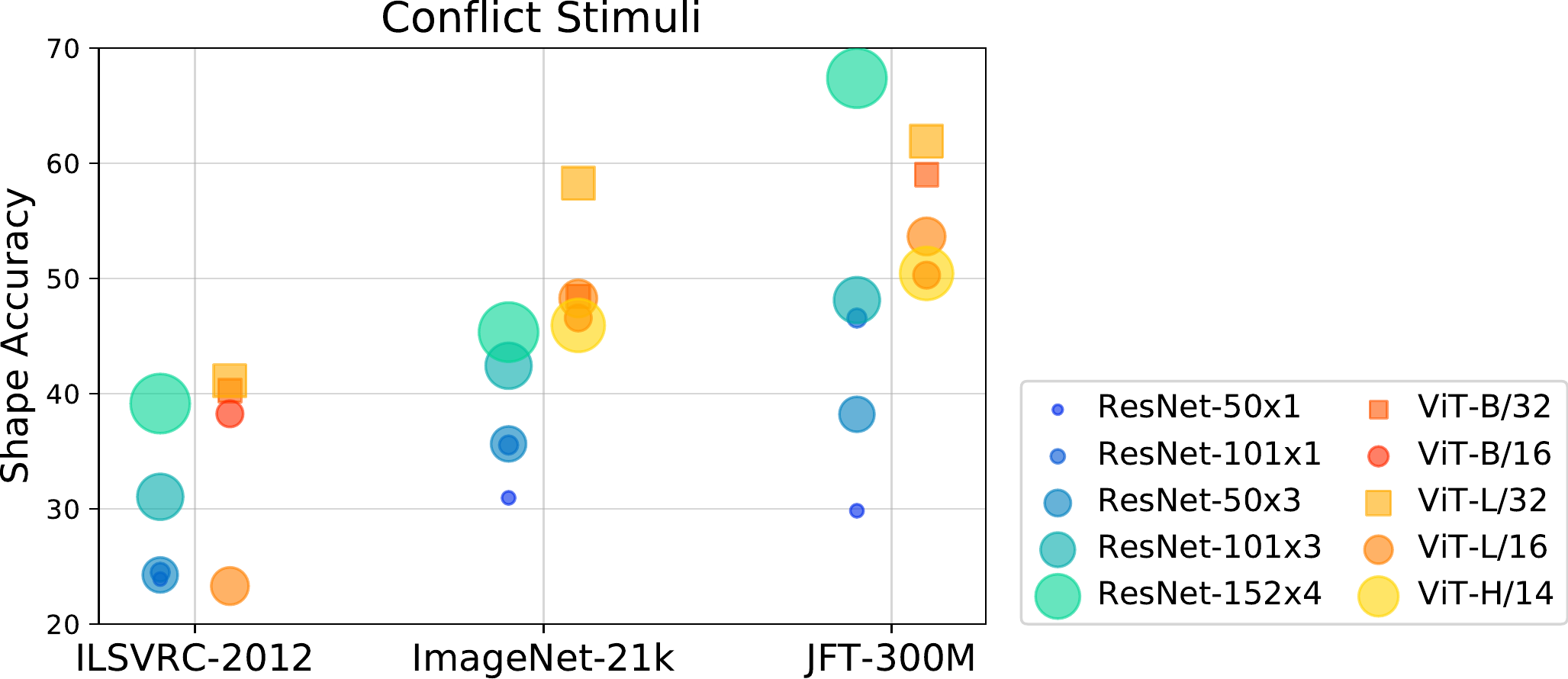}
\caption{\textbf{Texture and Shape}. Shape accuracy of ViT and \rn\ models on \cs\ \cite{geirhos2018imagenet}. In contrast to other robustness results, shape accuracy is more a function of patch size, than model size. $32 \times 32$ patch-sized ViT models do better than $16 \times 16$ ones.}\label{fig:geirhos_conflict_stimuli}
\end{figure}

The results are shown in Fig.~\ref{fig:geirhos_conflict_stimuli}.
An interesting observation is that the larger patch-sized ($32 \times 32$) ViT models perform better than the smaller patch-sized ($16 \times 16$) variants. This trend is different from what we see for clean accuracy as well as for \ic\ \ir\ and \ia. This may be due to larger patch inputs preserving object shape more than the smaller patches. We also observe that unlike in all other experiments, the performance of \rn{}s trained on \jft\ is not ordered by model size. 

\begin{figure}[!ht]
\centering
\includegraphics[width=\columnwidth]{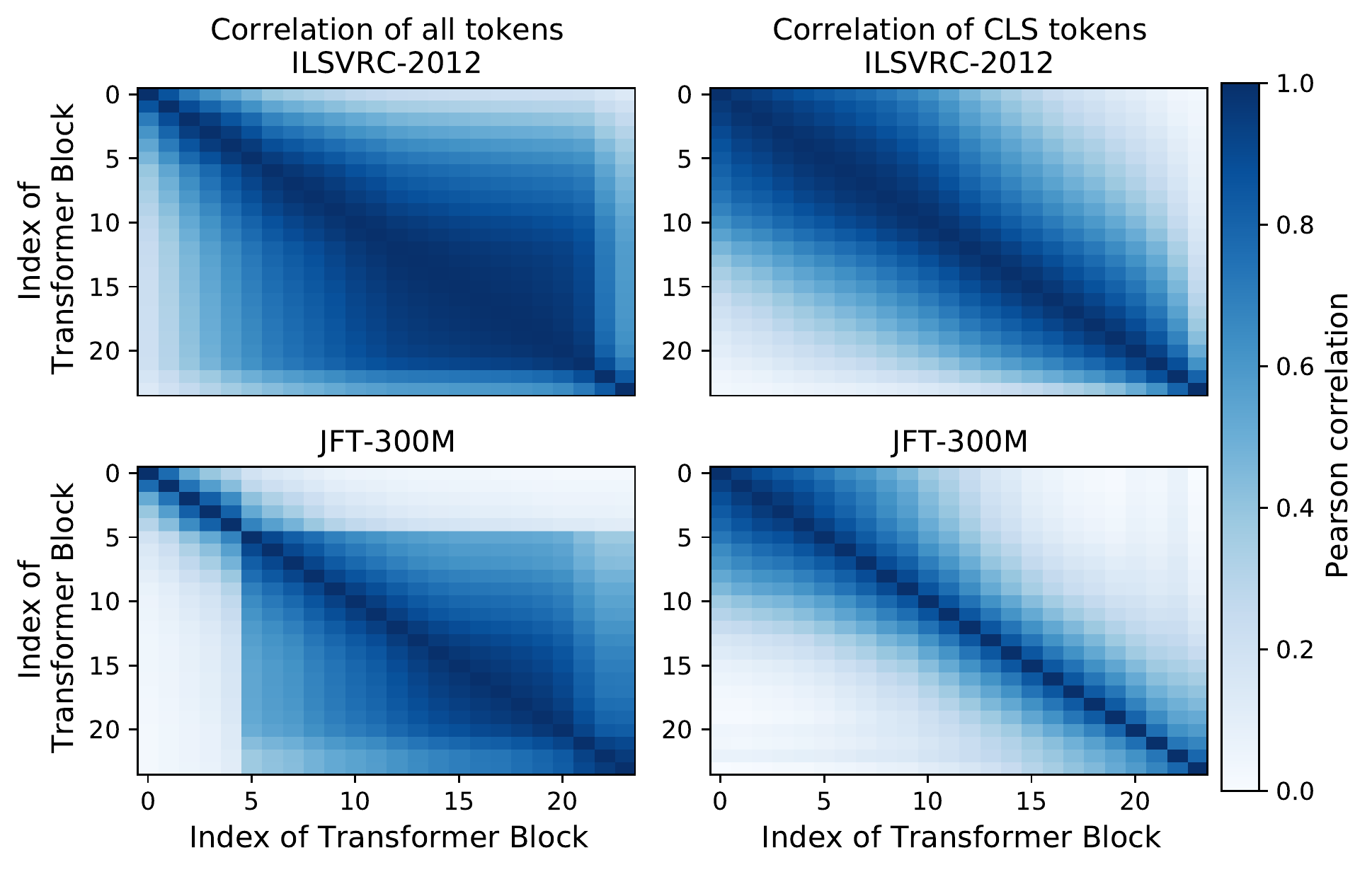}
\caption{\textbf{Representation Correlation Study}. We compare the representations (hidden features) after each Transformer block to those of all other blocks. When taking all tokens into account \emph{(left)}, we observe that representations are increasingly correlated towards the end of the network. In contrast, when only looking at the CLS token \emph{(right)}, the representations become less correlated throughout the network. A potential explanation could be that early layers focus on interactions among spatial tokens whereas later layers focus on the interactions between spatial tokens and the CLS token. Additional results can be found in Appendix~\ref{appendix:layer_correlation}. \label{fig:activation_study}}
\end{figure}

\begin{figure*}[h]
\centering
\includegraphics[width=\textwidth]{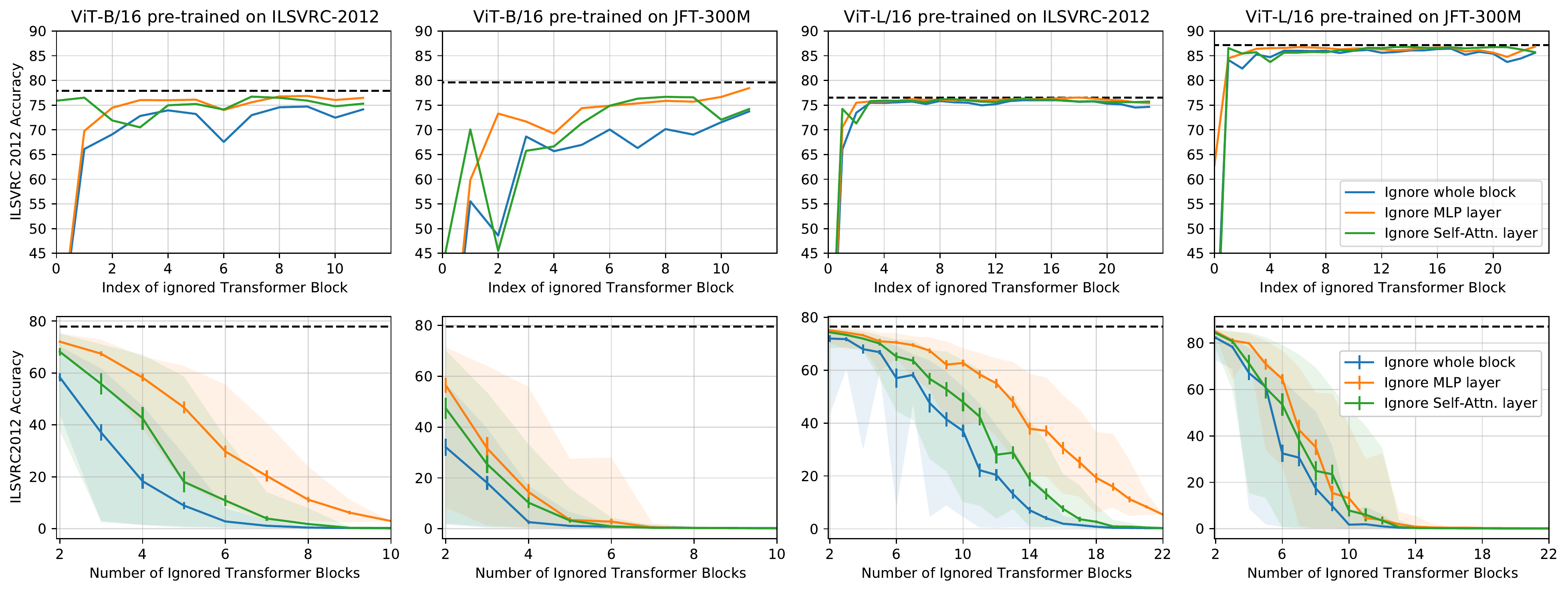}
\caption{\textbf{Lesion Study}. \emph{(top)} Evaluation of ViT models when individual blocks are removed from the model after training. We notice that besides the first block, one can remove any single block, self-attention or MLP from the trained model without substantially degrading performance. The larger models and the models trained only on ILSVRC-2012 are less impacted by the removal of individual layers. \emph{(bottom)} Evaluation of ViT models when $n > 1$ random blocks are removed from the model after training, while always keeping the first block. 
For each $n$, results are from 25 independent samples of $n$ blocks and we show the average accuracy (line), standard error (error bars) and min/max (shaded area) across samples.
We observe that there is less redundancy in smaller models and their accuracy drops quickly with removal of few blocks. Interestingly, we observe that removing self-attention layers hurts more than removing MLP layers. Additional results can be found in Appendix~\ref{appendix:lesion}. \label{fig:lesion_study}}
\end{figure*}

\section{Robustness to Model Perturbations} \label{sec:robustness_to_model_perturbations}
In this section we present our experiments on understanding the information flow in ViT models, by computing layer correlations, lesion studies and restricting attention. We first study how representation of input patches evolves in the ViTs by computing their block level correlations.

\paragraph{Layer correlations} We compute correlations between the representations of each Transformer block with the rest. In the left plots of Fig.\,\Ref{fig:activation_study} we present the correlations between representations of all blocks on ViT-L/16 for 2 datasets. Additional results for different models/datasets can be found in Appendix~\ref{appendix:layer_correlation}. We first notice that representation from many blocks towards later layers appear to be highly correlated, indicating a large amount of redundancy. Specifically, we observe that the layers organize into larger groups. In fact a similar pattern can be observed in \rn{}s, where downsampling layers separate the model into groups with different spatial resolutions. Surprisingly, despite lacking such inductive bias, ViT models also appear to organize layers into stages---the most
striking example being a very large, highly correlated group formed by the later layers, where representations appear to change only slightly.

Recall that ViT models append a special CLS token into the input sequence, whose representation is used to make the final classification. We next look at the correlation of the CLS token representations. Looking at this token in isolation, we see a different pattern (see right side of Fig.\,\Ref{fig:activation_study}): the representation of CLS tokens only changes slowly at the beginning of the network, but changes rapidly during the later layers. This indicates that the later layers of the network only provide limited updates to the representations of the individual patches, but focus on consolidating the information needed for the classification in the CLS token.

\begin{figure*}[!t]
    \centering
    \includegraphics[width=\textwidth]{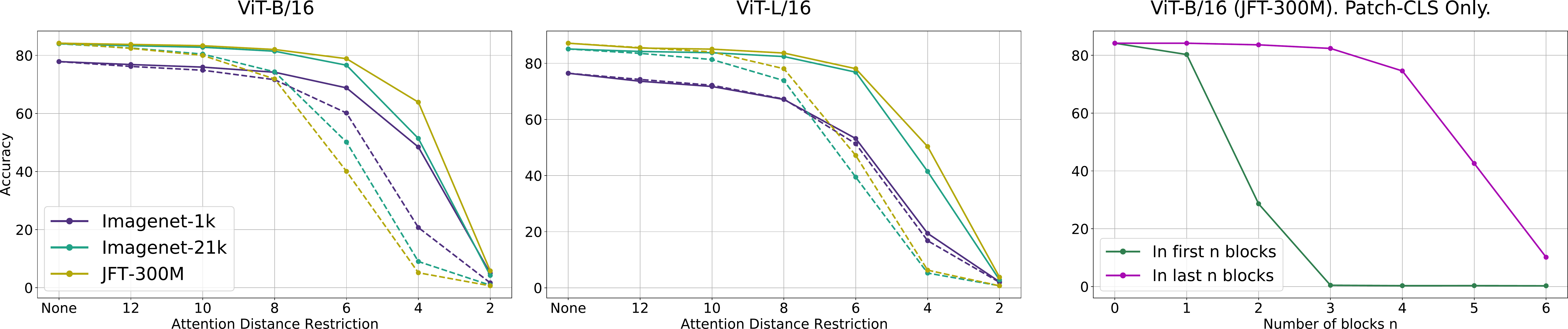}
    \caption{\textbf{Restricted Attention}. \emph{(Left, Center)} We evaluate two ViT models, pre-trained on different datasets, in terms of ILSVRC-2012 validation set accuracy when restricting attention among patches to only pairs that lie within a certain maximum horizontal or vertical distance. Dashed lines show the results with an equivalent amount of masking where pairs of patches to mask are chosen randomly (as a random permutation of the mask matrix used for spatially restricted attention). \emph{(Right)} For the ViT-B/16 model pre-trained with JFT-300M, we consider restricting attention to only between patches and the CLS token (without any attention between patches). We report accuracy applying this restriction to a subset of Transformer blocks---both at the beginning and at the end---of the model.}\label{fig:spat-ra}
\end{figure*}

\paragraph{Lesion study}
The presence of highly correlated representations across blocks raises the question whether the respective blocks are redundant. Previous works~\cite{veit2016residual, greff2017highway} have shown that layers in residual networks exhibit a large amount of redundancy, and that almost any individual layer can be removed after training without hurting performance. 
Following that line of work, we perform a lesion study on ViTs where we remove single blocks from an already trained network during inference, such that information has to flow through the skip connection. 
Each block contains two skip connections, and we separately investigate the effects of deleting MLP, self-attention layers, or the whole block.
This approach is similar to~\cite{michel2019sixteen}, but for whole layers and including the MLP block.
As shown in the top row of Fig.\,\Ref{fig:lesion_study}, it indeed appears that besides the first block one can remove any single block from the model without substantially degrading performance. This is in line with results reported for ResNets.

We next investigate the effect of removing several layers, while always keeping the first block. We observe that as more layers get removed, the performance gradually deteriorates (bottom row of Fig\,\Ref{fig:lesion_study}), with larger models being more robust to layer removal. We also notice that the amount of training data also influences robustness: models pre-trained on large datasets are less robust to layer removal, indicating perhaps higher model utilization. %
The results further show that removing individual layers reduces accuracy less than removing full blocks, indicating that there is limited co-adaptation among the components within each Transformer block. Lastly, we notice that removing MLP layers hurts the model less than removing the same number of self-attention layers, indicating the relative importance of self-attention. 
This behavior seems to be different to transformer models in NLP, which as alluded to by~\cite{michel2019sixteen} might behave in the opposite way. We have also observed this phenomenon in our own experiments.
Additional result from our lesion study can be found in Appendix~\ref{appendix:lesion}.

\paragraph{Restricted attention}
Finally, we study the extent to which ViT models rely on long-range attention. 
We evaluate this by spatially
\emph{restricting} the attention between patches to those that lie within a certain distance.We apply this restriction during inference only, by passing a spatial distance-based mask for attention between patches. %
Note that the masks always allow attention between the CLS token and all patches.

Figure~\ref{fig:spat-ra} shows that even though these models were trained assuming unrestricted attention, they degrade gracefully when inter-patch attention is restricted to be local. We also compare to a baseline of \emph{randomly} restricting attention by the same amount---achieved by using a random (but fixed across experiments) permutation of the mask matrix. We see that in this case, the degradation in performance is significantly higher in most cases---with a notable exception being the large ViT model that was trained only on \ione\ data.
Our last evaluation in Fig.~\ref{fig:spat-ra} considers the extreme version of this case, when attention is only allowed between patches and the CLS token, but not among patches, for a subset of Transformer blocks in the network---applying it either only to blocks at the beginning or at the end. Interestingly, we find that removing inter-patch attention completely at the end of the network has relatively little effect on accuracy---although this is consistent with our earlier observation that in the final few blocks of the network, it is the CLS token that is primarily being updated. In contrast, disrupting inter-patch attention in the initial blocks causes a significant degradation in accuracy.

To summarize, we find that ViT models contain a surprising amount of redundancy, which indicates that the model could be heavily pruned during inference. %

\section{Takeaways}

In this paper, we studied different aspects of robustness in ViT models, making a number of observations. Some of these confirmed existing intuitions about neural networks for vision, while others were perhaps surprising. We summarize they key takeaways from our analysis below:

\begin{itemize}[leftmargin=1em,itemsep=0pt,parsep=0pt,topsep=0pt,partopsep=0pt]
    \item Consistent with~\cite{dosovitskiy2020image}, we find that ViT models generally outperform \rn{}s and scale better with model size, \emph{when trained on sufficient data}. Crucially, the above is true also of robustness. We found that relative accuracy on the standard \ione\ validation set is predictive of performance under a diverse array of perturbations.
    \item We discovered that FGSM attacks fare better against ViT models than against \rn{}s. However, \rn\ models are not fundamentally more robust since both kinds of models are equally vulnerable to perturbations computed using PGD (which is more successful than the simpler FGSM). However, the optimal perturbations for the two kinds of models are very different and do not transfer.
    \item We found that choice of patch size in ViT models plays a role in their robustness. Smaller patch sizes make ViT models more robust to adversarial spatial transformations, but also incresae their texture bias.
    \item Through correlation analysis, we discovered that ViT models organize themselves into correlated groups much like \rn{}s, despite having no explicit downsampling-based groups like \rn{}s. This analysis also showed that most updates in the later layers are to the representation of the CLS token, rather than to those of individual patches. Moreover, preventing attention between patches in later layers led to a relatively lower drop in accuracy.
    \item We also found that despite their ability to allow patches to communicate globally, restricting attention to be local has a relatively lower effect on accuracy.
    \item Finally, our lesion studies showed that ViT models are fairly robust to removing individual layers. But contrary to observations on language tasks, we found that ViT models are more robust to the removal of MLP layers than self-attention ones.
\end{itemize}

\paragraph{Acknowledgements.}
We thank the authors of~\cite{dosovitskiy2020image} and of~\cite{kolesnikov2019big} for kindly sharing checkpoints of their pre-trained ViT and \rn\ models, respectively. 

{\small
\bibliographystyle{ieee_fullname}
\bibliography{references}
}

\onecolumn
\appendix
\pagenumbering{roman}
\begin{center}
{\large \bf Appendix}
\end{center}
~\\~
\section{Experimental Setup}

\renewcommand{\paragraph}[1]{\vspace{-0.5em}~\\\textbf{#1}.\ }

\paragraph{Image Preprocessing}
Following~\cite{kolesnikov2019big, dosovitskiy2020image}, we directly resize images---ignoring aspect ratio and without cropping---to the dimension expected by each network as input, normalizing intensities to the appropriate range. This size is $384 \times 384$ for most models.

\paragraph{Adversarial Perturbations} 
For both FGSM and PGD, we compute the gradients with respect to input pixels of the cross-entropy loss against the correct image label, and then use the sign of these gradients to update the images. We clip the updated image intensities to lie within the valid range after each update. The step sizes and overall $L_\infty$ norm of one gray level are translated to the expected intensity normalization of each model. 

\paragraph{Spatial Adversarial Attacks}
We provide additional details for the spatial adversarial grid attack that we employ in Sec.~\ref{sec:robust-spatial-advers-pert}. We perturb each image with a discrete set of spatial transformations. The attack is considered successful if any of the transformed images is incorrectly classified by the corresponding model model. The same fixed set of $2511 (9\times9\times31)$ transformations was used for all images and all models. The perturbation set corresponds to the vertices of a grid, which is defined by an outer product of sampled values of three parameters: horizontal translation, vertical translation, and rotation. The samples are equally spaced within each parameter's range. Engstrom et al.~\cite{engstrom2019exploring} use 5 values each for horizontal and vertical translation. We use a denser set of 9 values, so as to explore the space of translations at a finer resolution, and a range of $[-16, 16]$ pixels as the translation ranges, so as to span the largest patch size ($32 \times 32$) used by any of the ViT models. For rotations, we follow~\cite{engstrom2019exploring} using 31 values in the range $[-30^{\circ}, 30^{\circ}]$. When rotating and translating the images, we use bilinear interpolation and fill regions that lie outside the bounds of the original image with zeros (black pixels). 

\paragraph{Restricted Attention}
We pass masks to the Transformer attention layers to evaluate the effect of restricted attention. After the initial embedding layer, the image is transformed to a flat sequence of patches and a CLS token. To compute the pair-wise mask between all entries of that sequence, we consider the spatial location of the patches, and mask out pairs whose distance (on the patch grid, which is of size $384/16 \times 384/16$ for the models we consider) along the $x-$ or $y-$ axis is greater than the restricted distance. Our mask always allows attention between the CLS token and all patches.

\FloatBarrier  %

\section{Raw Accuracy Values}
Table~\ref{table:raw_accuracy} shows the raw accuracy values corresponding to the results shown in Figs.~\ref{fig:imagenet-c-r-a-cs} and \ref{fig:geirhos_conflict_stimuli}.

\begin{table*}[hbt!]
\centering
\begin{tabular}{lccccc}
\toprule
{} &  \bf ILSVRC-2012 & \bf ImageNet-C & \bf ImageNet-R & \bf ImageNet-A & \bf Conflict Stimuli \\
\midrule

\multicolumn{5}{l}{\em Models trained on \ione}\\
ResNet-50x1  &       76.80\% &      46.14\% &      21.45\% &       4.15\% &            23.91\% \\
ResNet-101x1 &       78.00\% &      50.24\% &      23.00\% &       6.28\% &            24.49\% \\
ViT-B/32     &       73.37\% &      48.77\% &      20.23\% &       3.80\% &            40.28\% \\
ViT-B/16     &       77.91\% &      52.22\% &      21.90\% &       7.00\% &            38.26\% \\
ResNet-50x3  &       80.00\% &      51.09\% &      23.62\% &       7.15\% &            24.27\% \\
ViT-L/32     &       71.18\% &      47.73\% &      19.14\% &       3.31\% &            41.13\% \\
ViT-L/16     &       76.50\% &      49.31\% &      17.87\% &       6.68\% &            23.30\% \\
ResNet-101x3 &       80.30\% &      53.36\% &      24.47\% &       9.07\% &            31.05\% \\
ResNet-152x4 &       80.40\% &      54.46\% &      25.82\% &      11.64\% &            39.14\% \vspace{0.25em}\\

\multicolumn{5}{l}{\em Models trained on \itwentyone}\\
ResNet-50x1  &       80.10\% &      52.97\% &      26.82\% &       9.92\% &            30.96\% \\
ResNet-101x1 &       82.40\% &      58.08\% &      30.37\% &      15.73\% &            35.54\% \\
ViT-B/32     &       81.27\% &      62.79\% &      31.79\% &      16.29\% &            48.45\% \\
ViT-B/16     &       83.98\% &      65.83\% &      37.99\% &      26.65\% &            46.58\% \\
ResNet-50x3  &       84.10\% &      59.87\% &      34.04\% &      20.40\% &            35.64\% \\
ViT-L/32     &       81.03\% &      65.05\% &      34.33\% &      19.04\% &            58.26\% \\
ViT-L/16     &       85.12\% &      70.03\% &      40.64\% &      28.12\% &            48.27\% \\
ResNet-101x3 &       84.50\% &      63.07\% &      35.93\% &      24.97\% &            42.42\% \\
ViT-H/14     &       85.11\% &      71.13\% &      39.99\% &      31.00\% &            45.93\% \\
ResNet-152x4 &       84.80\% &      65.31\% &      39.67\% &      30.01\% &            45.34\% \vspace{0.25em}\\

\multicolumn{5}{l}{\em Models trained on \jft}\\
ResNet-50x1  &       79.00\% &      51.26\% &      35.52\% &      11.35\% &            29.84\% \\
ResNet-101x1 &       81.60\% &      59.02\% &      45.45\% &      18.77\% &            46.57\% \\
ViT-B/32     &       80.72\% &      62.76\% &      43.17\% &      12.07\% &            58.99\% \\
ViT-B/16     &       84.14\% &      67.70\% &      53.02\% &      25.49\% &            50.29\% \\
ResNet-50x3  &       84.90\% &      64.16\% &      55.41\% &      31.15\% &            38.21\% \\
ViT-L/32     &       84.40\% &      70.94\% &      55.38\% &      22.81\% &            61.91\% \\
ViT-L/16     &       87.13\% &      76.08\% &      65.28\% &      44.00\% &            53.65\% \\
ResNet-101x3 &       86.30\% &      69.15\% &      62.64\% &      40.21\% &            48.12\% \\
ViT-H/14     &       88.33\% &      78.74\% &      69.91\% &      55.85\% &            50.44\% \\
ResNet-152x4 &       87.30\% &      73.41\% &      68.52\% &      51.23\% &            67.39\% \\

\bottomrule
\end{tabular}\vspace{0.5em}
\caption{\textbf{Raw Accuracy.} Accuracy of ViT and \rn\ models on different datasets. \ic~\cite{hendrycks2019robustness}, \ir~\cite{hendrycks2020many}, and \ia~\cite{hendrycks2019natural}\ are designed to evaluate robustness in the presence of ``natural corruptions", ``naturally occurring distribution shifts", and ``natural adversarial examples", respectively. \cs~\cite{geirhos2018imagenet} is designed to evaluate the degree to which a model is biased towards relying on texture over shape for image classification. For \ic\ the accuracy is averaged across all corruption types (including corruptions in the `extra' group), and over all severity levels.}\label{table:raw_accuracy}
\end{table*}

\section{\ic\ Detailed Results}\label{appendix:detailed_results}
The \ic\ benchmark~\cite{hendrycks2019robustness} includes 15 types of synthetically generated corruptions, grouped into 4 categories: `noise', `blur', `weather' and `digital'. Each corruption type has five levels of severity, resulting in 75 distinct corruptions. The benchmark also includes an `extra' group with additional corruptions. The results in Fig.~\ref{fig:imagenet-c-r-a-cs} are averaged across 95 distinct corruptions (75 from the corruption groups, and 20 from the `extra' group). In this section we provide more detailed results.

\begin{figure*}[hbt!]
\includegraphics[width=\textwidth]{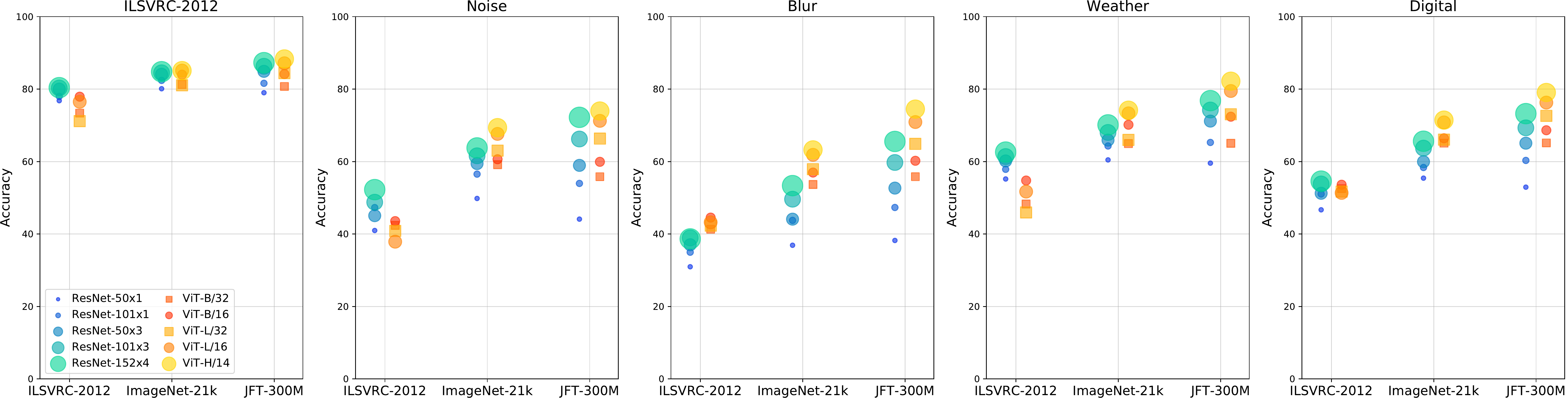}
\caption{\textbf{ImageNet-C Corruption Groups}. Accuracy of ViT and \rn\ models on \ione\ (clean) and the different corruption groups in \ic. The accuracy for each group is averaged across all corruption types in the group, and over all severity levels.}\label{fig:imagenet-c_corruption_groups}
\end{figure*}

\paragraph{Corruption Groups}
In Fig.~\ref{fig:imagenet-c_corruption_groups} we show accuracy for each of the corruption groups: `noise', `blur', `weather' and `digital'. The results for each corruption group are averaged across all corruption types in the group, and over all severity levels.

\FloatBarrier  %

\begin{figure*}[!h] 
\includegraphics[width=\textwidth]{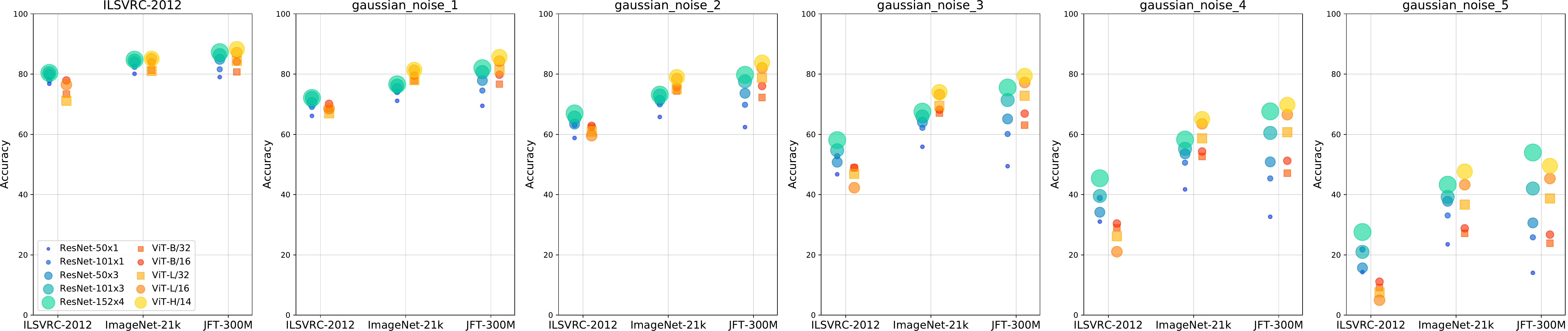}
\caption{\textbf{ImageNet-C `gaussian noise'}. Accuracy of ViT and \rn\ models on \ione\ (clean) and `gaussian noise'.}
\label{fig:imagenet-gaussian_noise}
\vspace{-0.1in}
\end{figure*}

\begin{figure*}[!h] 
\includegraphics[width=\textwidth]{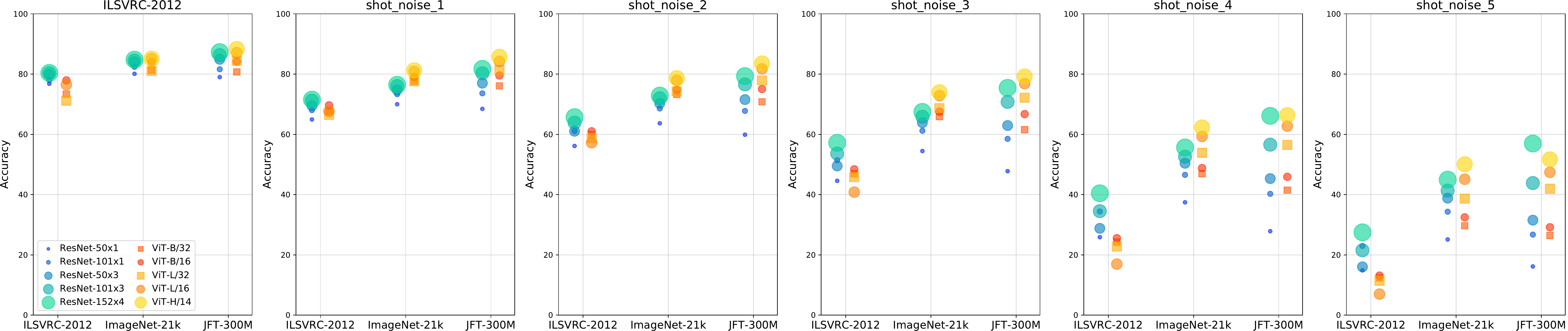}
\caption{\textbf{ImageNet-C `shot noise'}. Accuracy of ViT and \rn\ models on \ione\ (clean) and `shot noise'.}
\label{fig:imagenet-shot_noise}
\vspace{-0.1in}
\end{figure*}

\begin{figure*}[!h] 
\includegraphics[width=\textwidth]{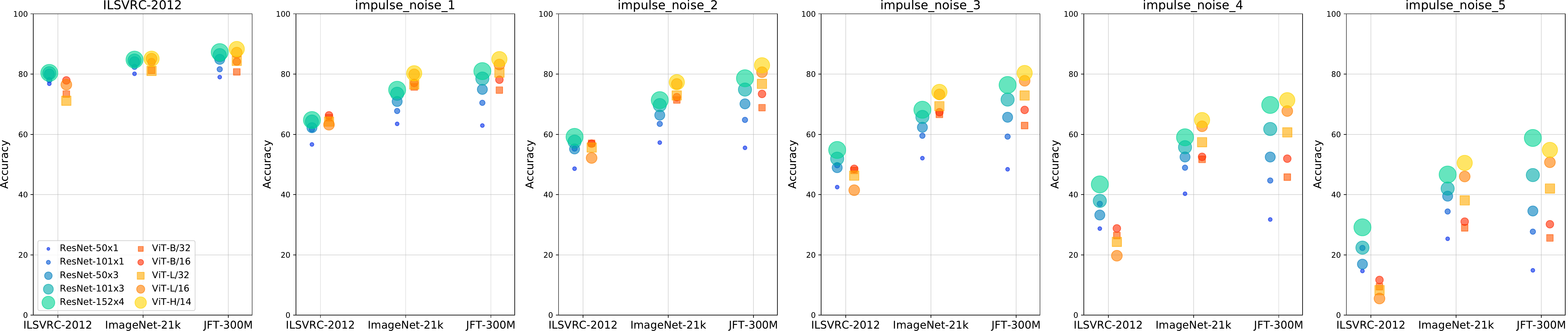}
\caption{\textbf{ImageNet-C `impulse noise'}. Accuracy of ViT and \rn\ models on \ione\ (clean) and `impulse noise'.}
\label{fig:imagenet-impulse_noise}
\vspace{-0.1in}
\end{figure*}

\begin{figure*}[!h] 
\includegraphics[width=\textwidth]{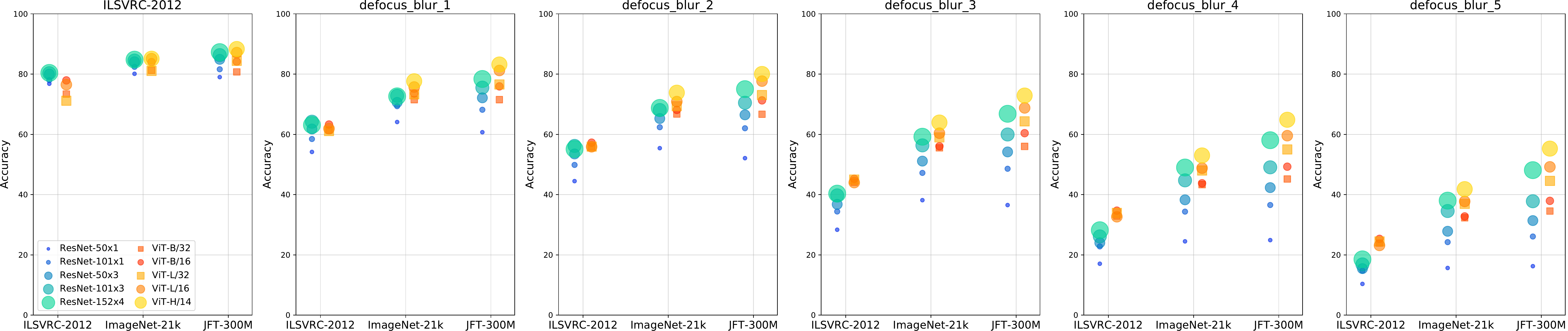}
\caption{\textbf{ImageNet-C `defocus blur'}. Accuracy of ViT and \rn\ models on \ione\ (clean) and `defocus blur'.}
\label{fig:imagenet-defocus_blur}
\vspace{-0.1in}
\end{figure*}

\begin{figure*}[!h] 
\includegraphics[width=\textwidth]{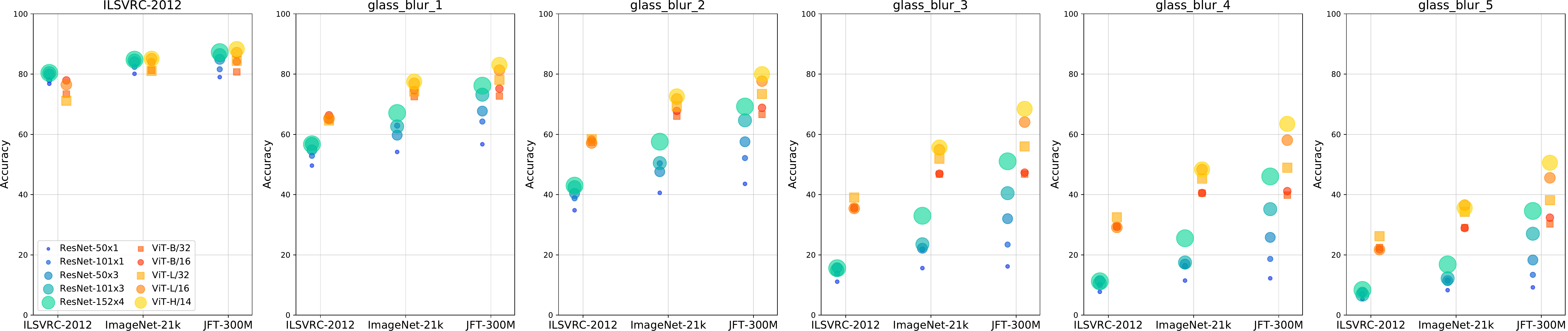}
\caption{\textbf{ImageNet-C `glass blur'}. Accuracy of ViT and \rn\ models on \ione\ (clean) and `glass blur'.}
\label{fig:imagenet-glass_blur}
\vspace{-0.1in}
\end{figure*}

\begin{figure*}[!h] 
\includegraphics[width=\textwidth]{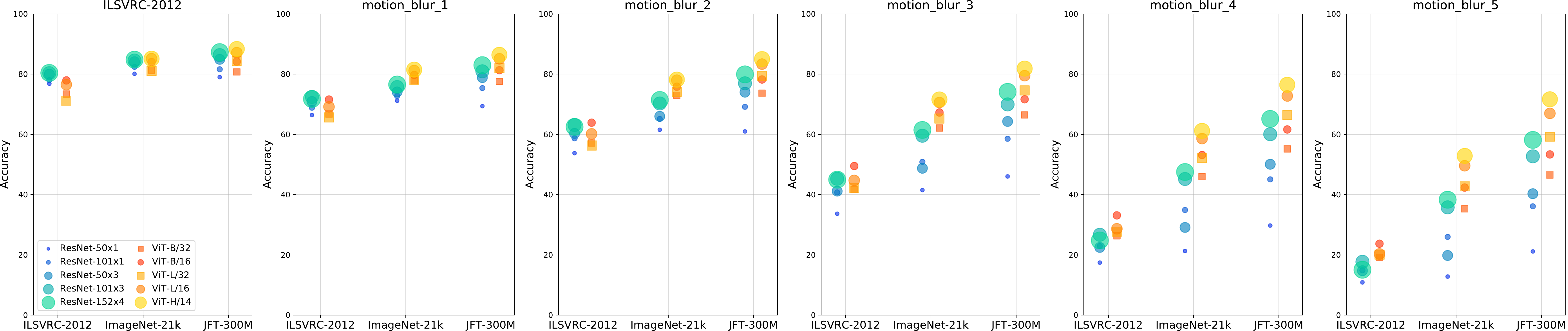}
\caption{\textbf{ImageNet-C `motion blur'}. Accuracy of ViT and \rn\ models on \ione\ (clean) and `motion blur'.}
\label{fig:imagenet-motion_blur}
\vspace{-0.1in}
\end{figure*}

\begin{figure*}[!h] 
\includegraphics[width=\textwidth]{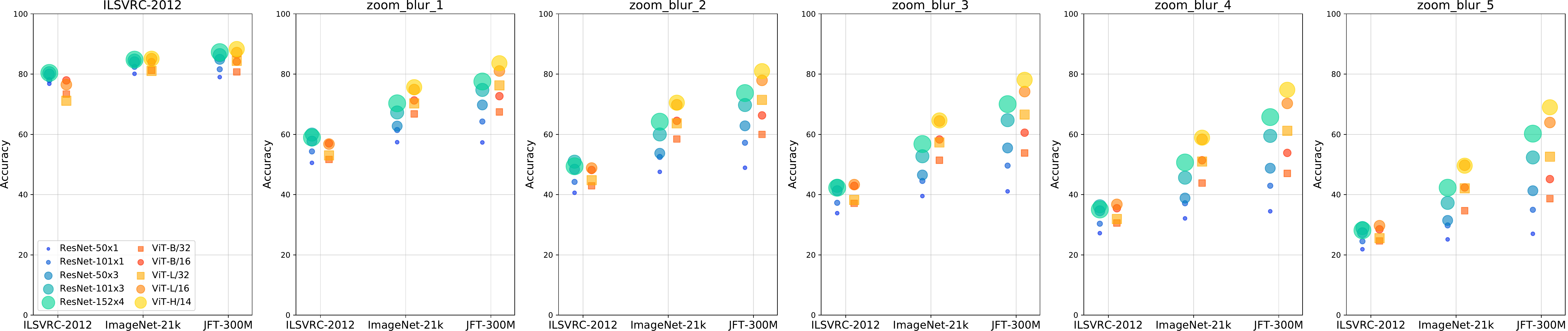}
\caption{\textbf{ImageNet-C `zoom blur'}. Accuracy of ViT and \rn\ models on \ione\ (clean) and `zoom blur'.}
\label{fig:imagenet-zoom_blur}
\vspace{-0.1in}
\end{figure*}

\begin{figure*}[!h] 
\includegraphics[width=\textwidth]{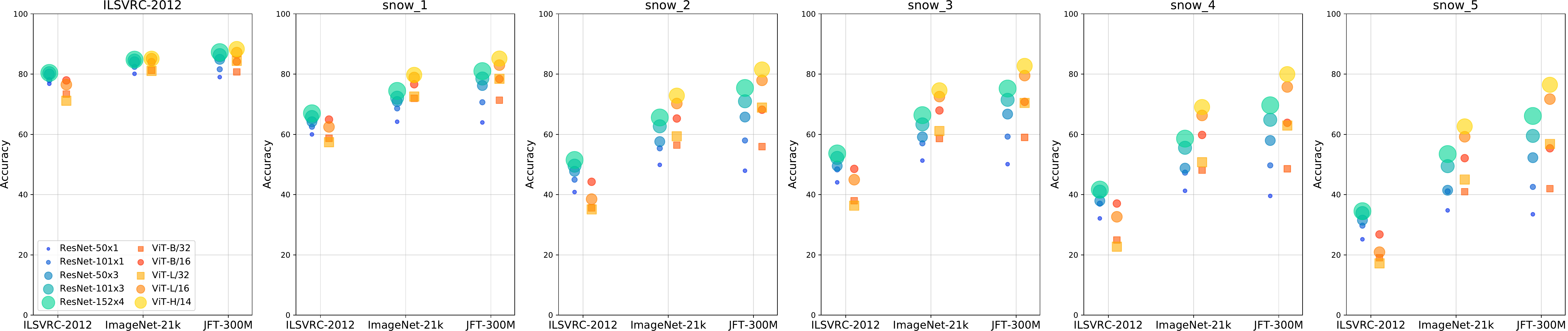}
\caption{\textbf{ImageNet-C `snow'}. Accuracy of ViT and \rn\ models on \ione\ (clean) and `snow'.}
\label{fig:imagenet-snow}
\vspace{-0.1in}
\end{figure*}

\begin{figure*}[!h] 
\includegraphics[width=\textwidth]{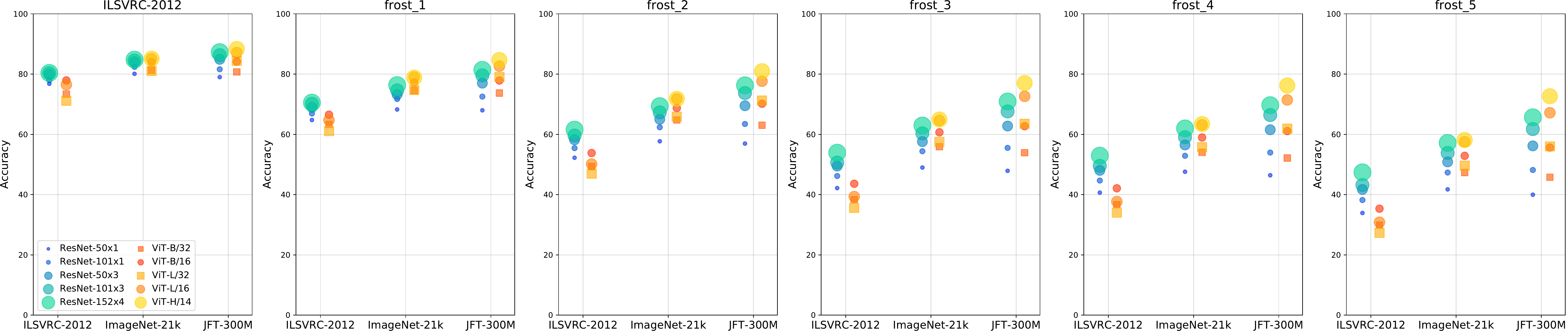}
\caption{\textbf{ImageNet-C `frost'}. Accuracy of ViT and \rn\ models on \ione\ (clean) and `frost'.}
\label{fig:imagenet-frost}
\vspace{-0.1in}
\end{figure*}

\begin{figure*}[!h] 
\includegraphics[width=\textwidth]{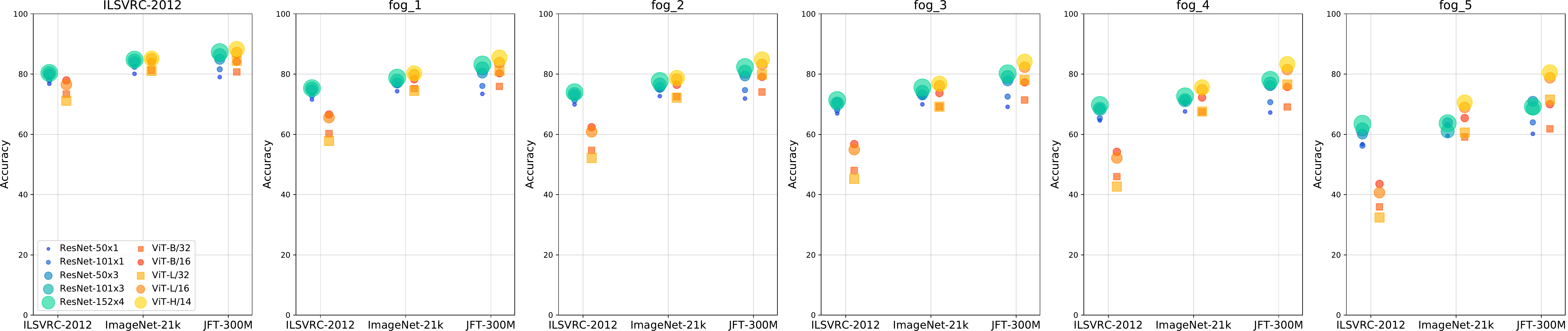}
\caption{\textbf{ImageNet-C `fog'}. Accuracy of ViT and \rn\ models on \ione\ (clean) and `fog'.}
\label{fig:imagenet-fog}
\vspace{-0.1in}
\end{figure*}

\begin{figure*}[!h] 
\includegraphics[width=\textwidth]{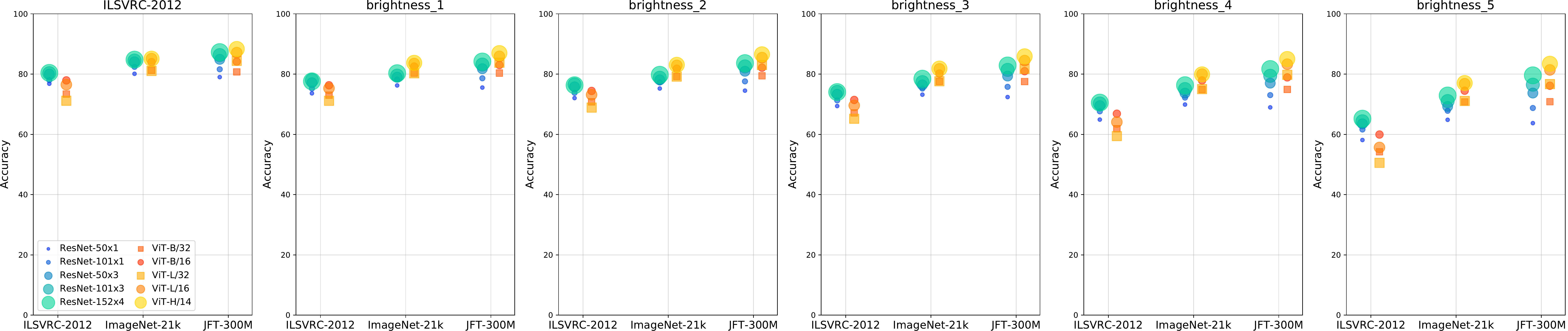}
\caption{\textbf{ImageNet-C `brightness'}. Accuracy of ViT and \rn\ models on \ione\ (clean) and `brightness'.}
\label{fig:imagenet-brightness}
\vspace{-0.1in}
\end{figure*}

\begin{figure*}[!h] 
\includegraphics[width=\textwidth]{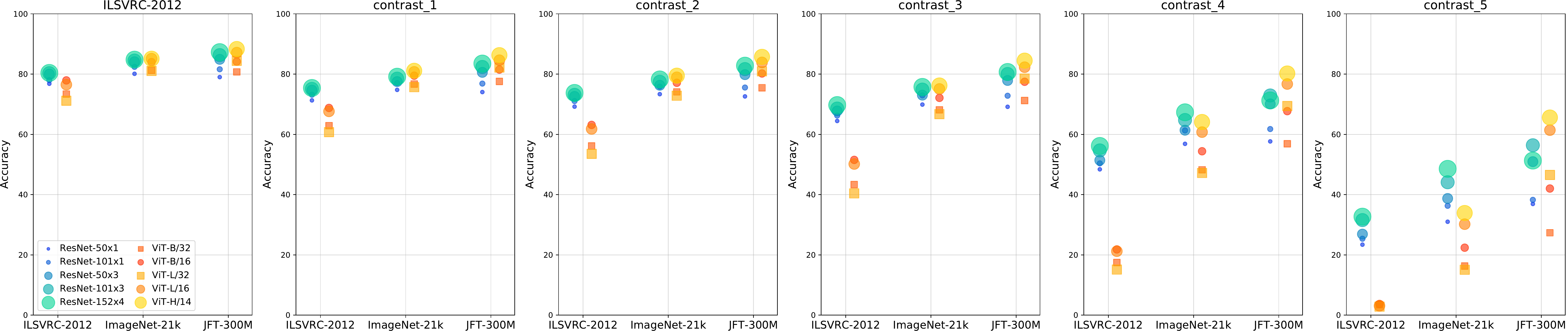}
\caption{\textbf{ImageNet-C `contrast'}. Accuracy of ViT and \rn\ models on \ione\ (clean) and `contrast'.}
\label{fig:imagenet-contrast}
\vspace{-0.1in}
\end{figure*}

\begin{figure*}[!h] 
\includegraphics[width=\textwidth]{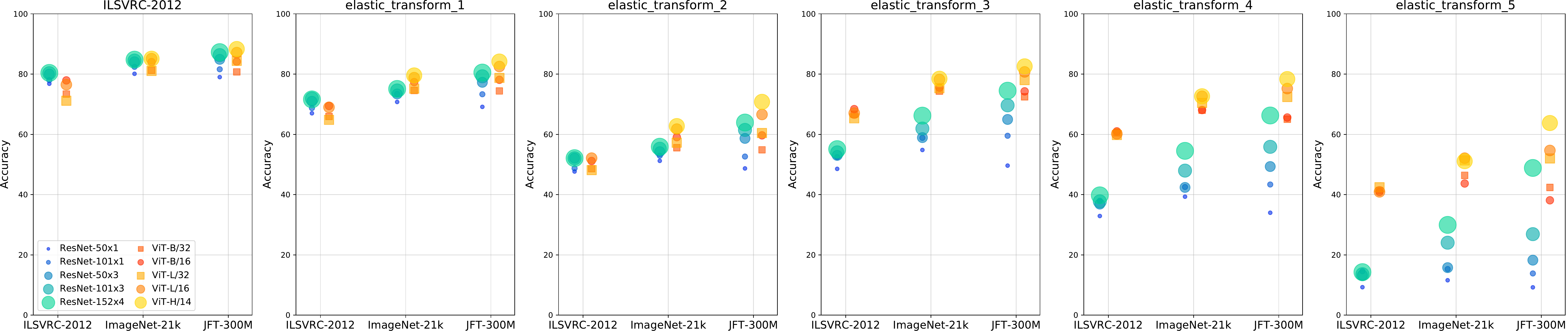}
\caption{\textbf{ImageNet-C `elastic transform'}. Accuracy of ViT and \rn\ models on \ione\ (clean) and `elastic transform'.}
\label{fig:imagenet-elastic_transform}
\vspace{-0.1in}
\end{figure*}

\begin{figure*}[!h] 
\includegraphics[width=\textwidth]{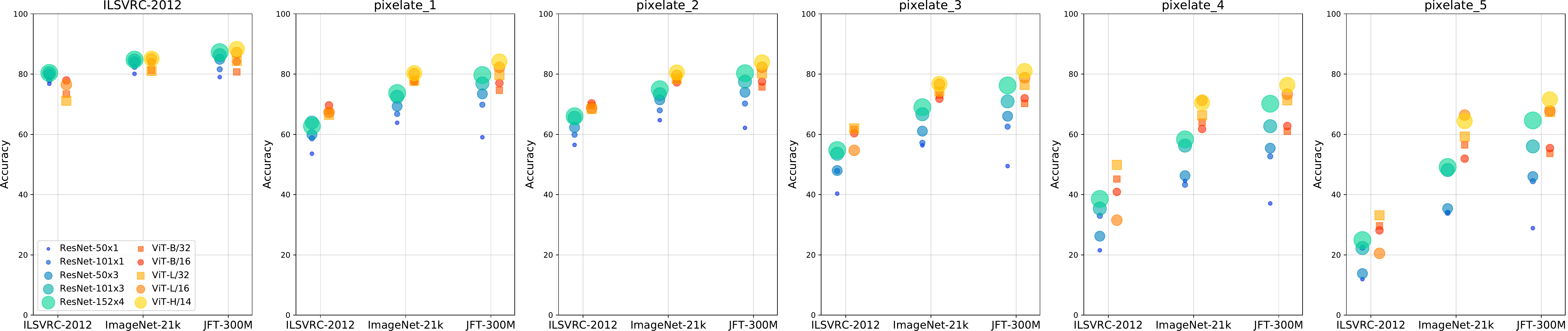}
\caption{\textbf{ImageNet-C `pixelate'}. Accuracy of ViT and \rn\ models on \ione\ (clean) and `pixelate'.}
\label{fig:imagenet-pixelate}
\vspace{-0.1in}
\end{figure*}

\begin{figure*}[!h] 
\includegraphics[width=\textwidth]{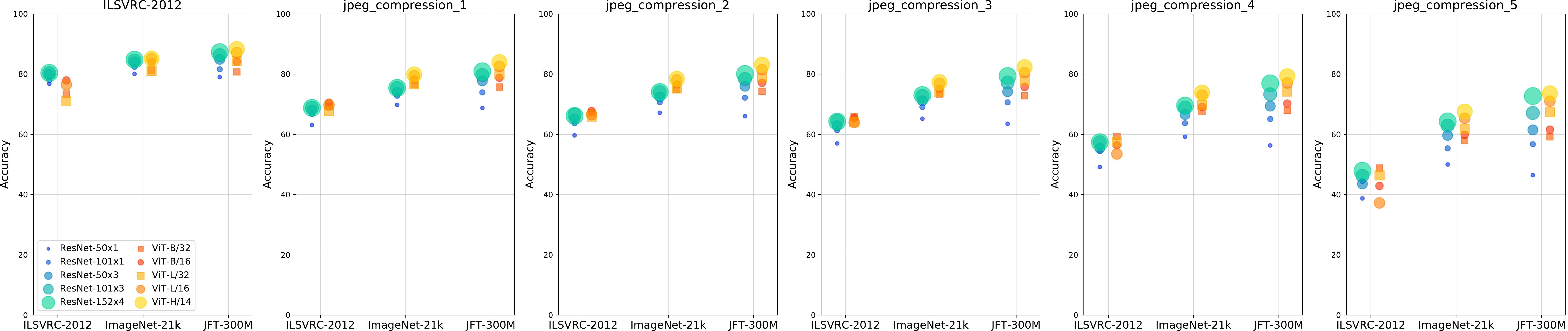}
\caption{\textbf{ImageNet-C `jpeg compression'}. Accuracy of ViT and \rn\ models on \ione\ (clean) and `jpeg compression'.}
\label{fig:imagenet-jpeg_compression}
\vspace{-0.1in}
\end{figure*}

\begin{figure*}[!h] 
\includegraphics[width=\textwidth]{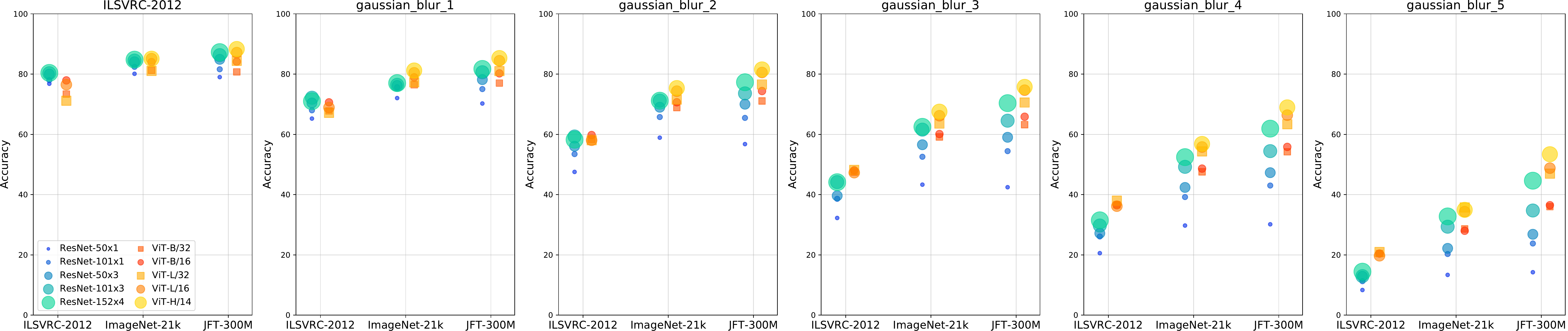}
\caption{\textbf{ImageNet-C `gaussian blur'}. Accuracy of ViT and \rn\ models on \ione\ (clean) and `gaussian blur'.}
\label{fig:imagenet-gaussian_blur}
\vspace{-0.1in}
\end{figure*}

\begin{figure*}[!h] 
\includegraphics[width=\textwidth]{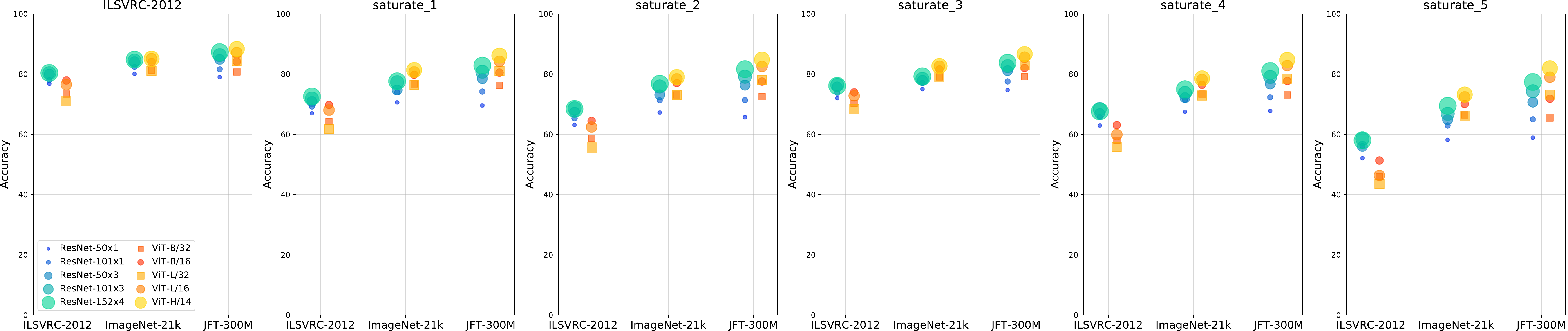}
\caption{\textbf{ImageNet-C `saturate'}. Accuracy of ViT and \rn\ models on \ione\ (clean) and `saturate'.}
\label{fig:imagenet-saturate}
\vspace{-0.1in}
\end{figure*}

\begin{figure*}[!h] 
\includegraphics[width=\textwidth]{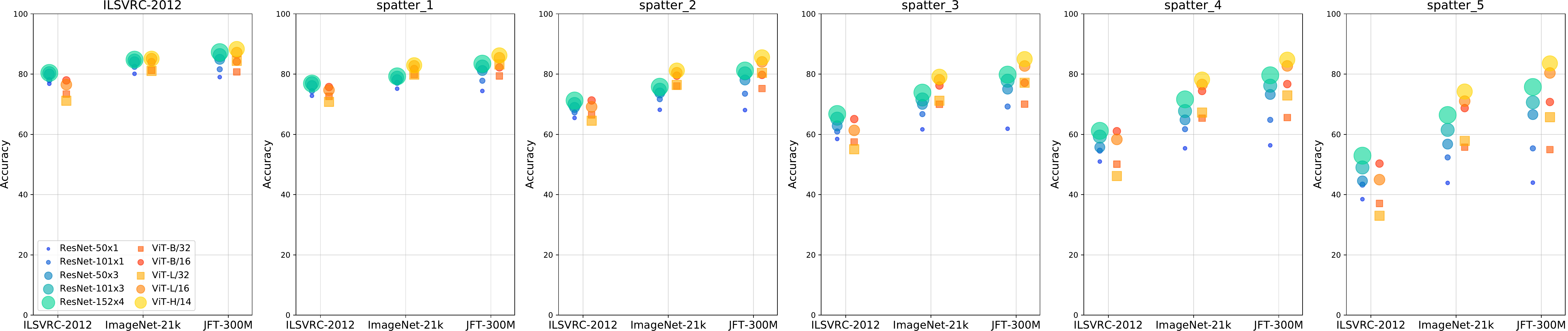}
\caption{\textbf{ImageNet-C `spatter'}. Accuracy of ViT and \rn\ models on \ione\ (clean) and `spatter'.}
\label{fig:imagenet-spatter}
\vspace{-0.1in}
\end{figure*}

\begin{figure*}[!h] 
\includegraphics[width=\textwidth]{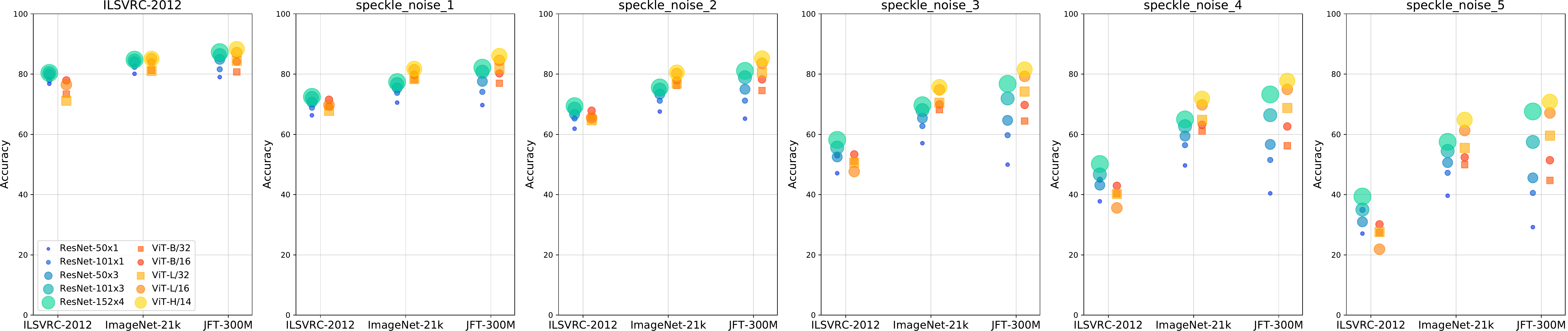}
\caption{\textbf{ImageNet-C `speckle noise'}. Accuracy of ViT and \rn\ models on \ione\ (clean) and `speckle noise'.}
\label{fig:imagenet-speckle_noise}
\vspace{-0.1in}
\end{figure*}

\paragraph{Individual Corruptions and Severities}
Next, Figures \ref{fig:imagenet-gaussian_noise}-\ref{fig:imagenet-speckle_noise} show detailed results for the 95 distinct corruptions in the \ic\ benchmark.

\FloatBarrier  %

\section{Robustness Scaling on \itwentyone}\label{appendix:i21k_scaling}

\begin{figure}[hbt!]
\centering
\includegraphics[width=0.55\columnwidth]{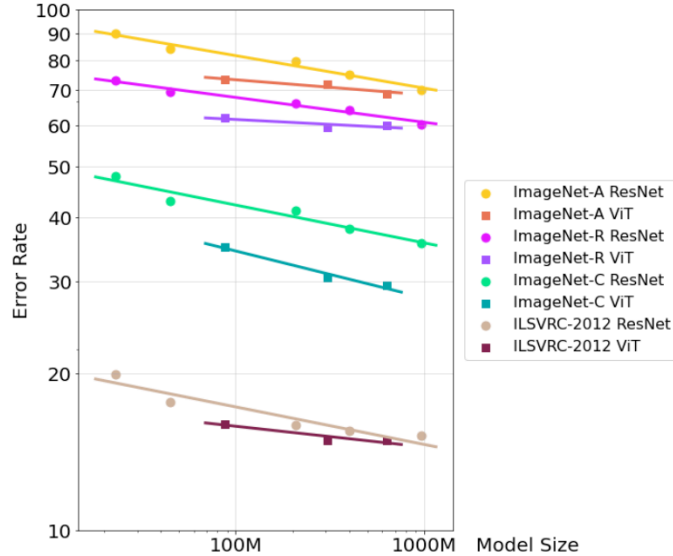}
\caption{\textbf{Performance of ViT and \rn\ Models on Different Datasets as a Function of the Number of Model Parameters}.  All models are pre-trained on ImageNet-21k and fine-tuned on \ione. We observe that this pretraining dataset is not large enough for the ViT models to exhibit better robustness scaling.}\label{fig:scaling_i21k}
\end{figure}

As we pointed out in Sec.~\ref{sec:robustness_to_input_perturbations}, the size of the pretraining dataset has a fundamental effect on the model's robustness, especially for ViTs. We compared the scaling of robustness on various benchmarks when all models are trained on ImageNet-21k in Fig.~\ref{fig:scaling_i21k}. With this smaller pretraining set, scaling up the ViT models does not offer better gains compared to scaling up ResNets in most cases, with ImageNet-C being an exception. 

\FloatBarrier  %

\section{Adversarial Perturbations: Accuracies with Self and Cross-Over Attacks}\label{appendix:adversarial-cross-over}
In Tables \ref{tab:supp-pgd} and \ref{tab:supp-fgsm}, we provide a full evaluation showing that adversarial perturbations computed using ViT models fail to cause incorrect outputs with ResNet models and vice-versa. For different variants of the ViT and ResNet models trained on different amounts of data, we report accuracies under adversarial perturbations---computed using PGD and FGSM---when evaluated using the models used to compute the perturbations vs.\ using a different model type (ViT or ResNet).

\begin{table}[!h]
\centering
\begin{tabular}{lcccccc}\toprule
& \bf ViT (clean) & \bf RN (clean) & \bf ViT$\rightarrow$ViT & \bf RN$\rightarrow$RN &  \bf ViT$\rightarrow$RN & \bf RN$\rightarrow$ViT\\\midrule

\multicolumn{5}{l}{\em Models trained on \ione}\\
ViT-B/16 vs.\ RN-101x1 & 77.8\% & 77.4\% &14.3\%  & 2.0\%  & 75.5\% & 77.4\%\\
ViT-B/32 vs.\ RN-50x3  & 73.0\% & 80.6\% &17.2\%  &  4.8\% & 79.0\% & 72.0\%\\
ViT-L/16 vs.\ RN-101x3 & 75.4\% & 80.2\% & 13.0\% &  7.3\% & 78.8\% & 74.1\%\\
ViT-L/32 vs.\ RN-152x4 & 71.5\% & 80.0\% & 16.0\% & 10.5\% & 79.1\% & 71.0\%\vspace{0.25em}\\

\multicolumn{5}{l}{\em Models trained on \itwentyone}\\
ViT-B/16 vs.\ RN-101x1 & 83.4\% & 82.1\% &10.3\%  &10.5\%  & 81.0\% & 82.9\%\\
ViT-B/32 vs.\ RN-50x3  & 81.1\% & 84.4\% &15.9\%  & 13.0\% & 83.7\% & 80.7\%\\
ViT-L/16 vs.\ RN-101x3 & 85.3\% & 85.8\% & 16.2\% & 14.4\% & 83.5\% & 84.7\%\\
ViT-L/32 vs.\ RN-152x4 & 81.9\% & 84.6\% & 24.7\% & 13.9\% & 84.0\% & 82.0\%\vspace{0.25em}\\

\multicolumn{5}{l}{\em Models trained on \jft}\\
ViT-B/16 vs.\ RN-101x1 & 85.6\% & 82.2\% & 5.4\%  & 8.1\%  & 79.7\% & 85.2\%\\
ViT-B/32 vs.\ RN-50x3  & 81.2\% & 83.9\% & 9.5\%  & 13.4\% & 82.2\% & 80.9\%\\
ViT-L/16 vs.\ RN-101x3 & 86.4\% & 86.1\% & 19.7\% & 19.5\% & 84.3\% & 85.8\%\\
ViT-L/32 vs.\ RN-152x4 & 86.7\% & 86.1\% & 17.6\% & 17.9\% & 85.4\% & 86.5\%\\\bottomrule
\end{tabular}\vspace{0.5em}
\caption{\textbf{Accuracy with PGD attacks.} We include the full set of results for adversarial perturbations applied to various ViT and ResNet (RN) models trained on different datasets. Shown here are accuracies on a subset of 1000 images of the \ione\ validation set (as used in Fig.~\ref{fig:adversarial} and Table \ref{tab:xfer})---on the original images as well as under adversarial perturbations computed using the ViT and ResNet models with PGD, when applied to the models used to compute the perturbations themselves, and when running inference on ResNet models using perturbations computed with ViT models and vice-versa.}\label{tab:supp-pgd}
\end{table}

\begin{table}[!h]
\centering
\begin{tabular}{lcccccc}\toprule
& \bf ViT (clean) & \bf RN (clean) & \bf ViT$\rightarrow$ViT & \bf RN$\rightarrow$RN &  \bf ViT$\rightarrow$RN & \bf RN$\rightarrow$ViT\\\midrule

\multicolumn{5}{l}{\em Models trained on \ione}\\
ViT-B/16 vs.\ RN-101x1 & 77.8\% & 77.4\% &30.6\%  &14.7\%  & 75.4\% & 77.2\%\\
ViT-B/32 vs.\ RN-50x3  & 73.0\% & 80.6\% &31.5\%  & 22.1\% & 78.5\% & 71.9\%\\
ViT-L/16 vs.\ RN-101x3 & 75.4\% & 80.2\% & 27.8\% & 23.6\% & 78.3\% & 73.7\%\\
ViT-L/32 vs.\ RN-152x4 & 71.5\% & 80.0\% & 26.3\% & 33.3\% & 78.1\% & 71.2\%\vspace{0.25em}\\

\multicolumn{5}{l}{\em Models trained on \itwentyone}\\
ViT-B/16 vs.\ RN-101x1 & 83.4\% & 82.1\% &31.3\%  &33.5\%  & 80.7\% & 82.7\%\\
ViT-B/32 vs.\ RN-50x3  & 81.1\% & 84.4\% &36.7\%  & 42.2\% & 83.3\% & 80.4\%\\
ViT-L/16 vs.\ RN-101x3 & 85.3\% & 85.8\% & 40.5\% & 44.4\% & 83.4\% & 85.1\%\\
ViT-L/32 vs.\ RN-152x4 & 81.9\% & 84.6\% & 41.6\% & 47.2\% & 84.2\% & 81.5\%\vspace{0.25em}\\

\multicolumn{5}{l}{\em Models trained on \jft}\\
ViT-B/16 vs.\ RN-101x1 & 85.6\% & 82.2\% &22.9\%  &30.6\%  & 79.9\% & 84.4\%\\
ViT-B/32 vs.\ RN-50x3  & 81.2\% & 83.9\% &25.7\%  & 40.0\% & 82.5\% & 80.4\%\\
ViT-L/16 vs.\ RN-101x3 & 86.4\% & 86.1\% & 43.2\% & 48.8\% & 83.4\% & 85.3\%\\
ViT-L/32 vs.\ RN-152x4 & 86.7\% & 86.1\% & 40.9\% & 52.7\% & 85.5\% & 85.7\%\\\bottomrule
\end{tabular}\vspace{0.5em}
\caption{\textbf{Accuracy with FGSM attacks.} Version of Table \ref{tab:supp-pgd} with perturbations computed using FGSM instead of PGD.}\label{tab:supp-fgsm}
\end{table}

\FloatBarrier  %

\section{Layer Correlation Analysis}\label{appendix:layer_correlation}
\begin{figure*}[hbt!]
    \centering
    \includegraphics[width=\textwidth]{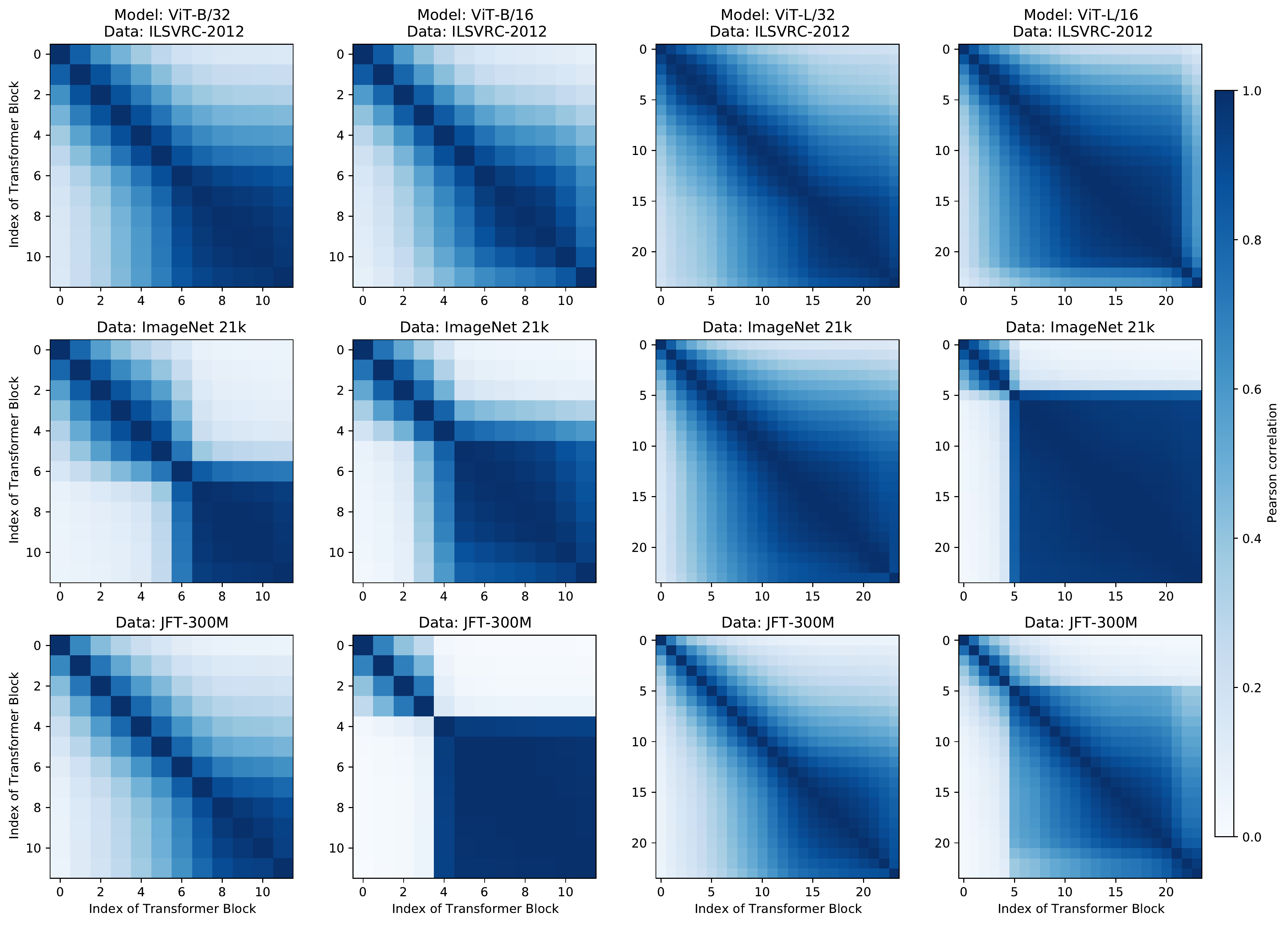}
    \caption{\textbf{Correlation of Representations Across Transformer Blocks}. We compare the representations (hidden features) after each Transformer block to those of all other blocks. We compare the similarity across representations using the absolute value of the Pearson correlation coefficient. We compare four different ViT models (ViT-B/32, ViT-B/16, ViT-L/32, ViT-L/16) and three different pre-training datasets (ILSVRC-2012, ImageNet 21k, JFT-300M). All models are fine-tuned on \ione\, and activations are calculated on a random subset of 4096 samples from the ILSVRC-2012 validation set. White: no correlation. Dark Blue: $|\rho|=1$.}
    \label{fig:correlation_study_appendix_all_tokens}
\end{figure*}

\begin{figure*}[hbt!]
    \centering
    \includegraphics[width=\textwidth]{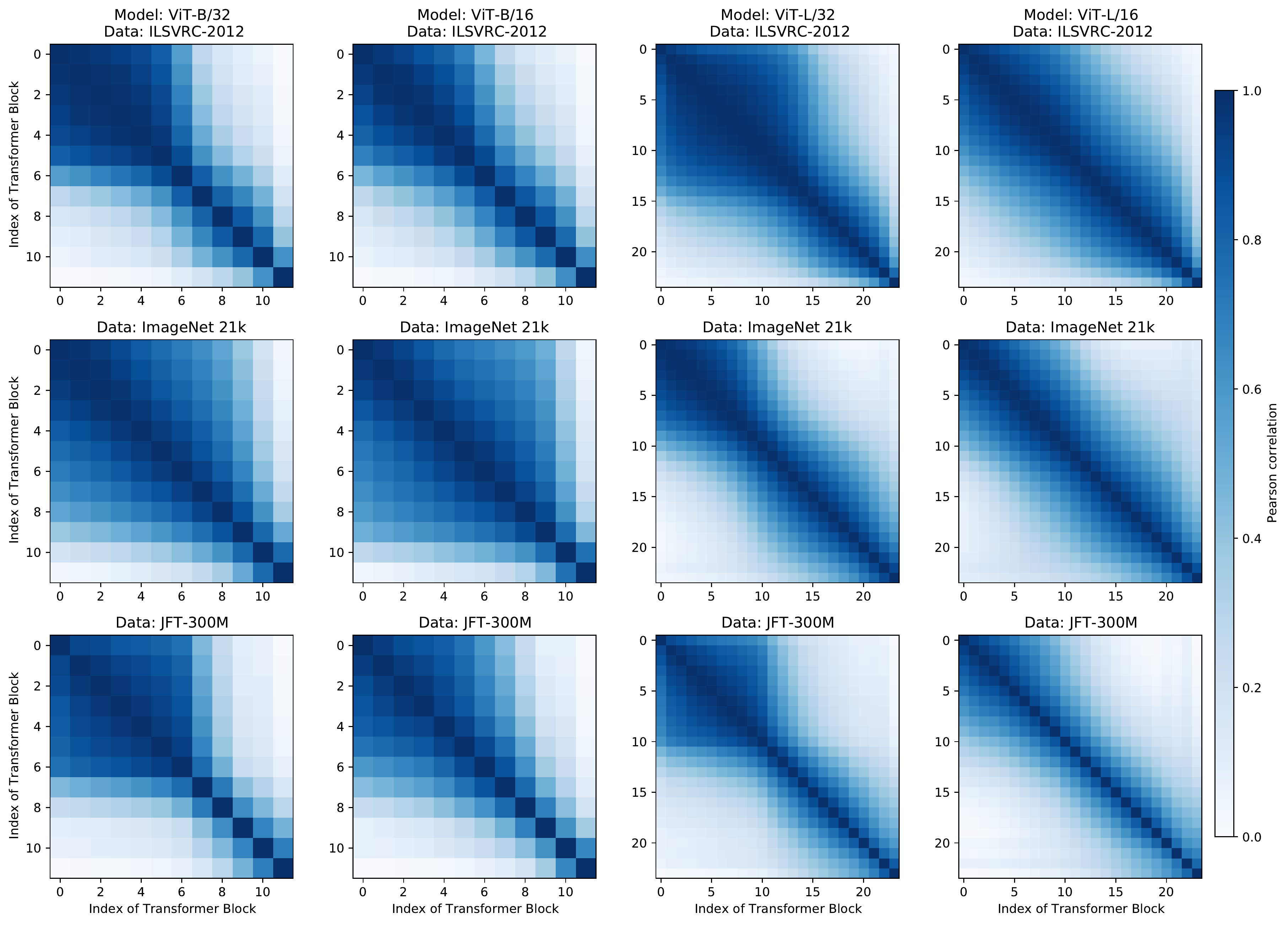}
    \caption{\textbf{Correlation of CLS Tokens Across Transformer Blocks}. We compare the representations (hidden features) of the CLS token after each Transformer block to those of all other blocks. We compare the similarity across representations using the absolute value of the Pearson correlation coefficient. We compare four different ViT models (ViT-B/32, ViT-B/16, ViT-L/32, ViT-L/16) and three different pre-training datasets (ILSVRC-2012, ImageNet 21k, JFT-300M). All models are fine-tuned on \ione\, and activations are calculated on a random subset of 4096 samples from the ILSVRC-2012 validation set. White: no correlation. Dark Blue: $|\rho|=1$.}
    \label{fig:correlation_study_appendix_cls_tokens}
\end{figure*}

\begin{figure*}[hbt!]
    \centering
    \includegraphics[width=\textwidth]{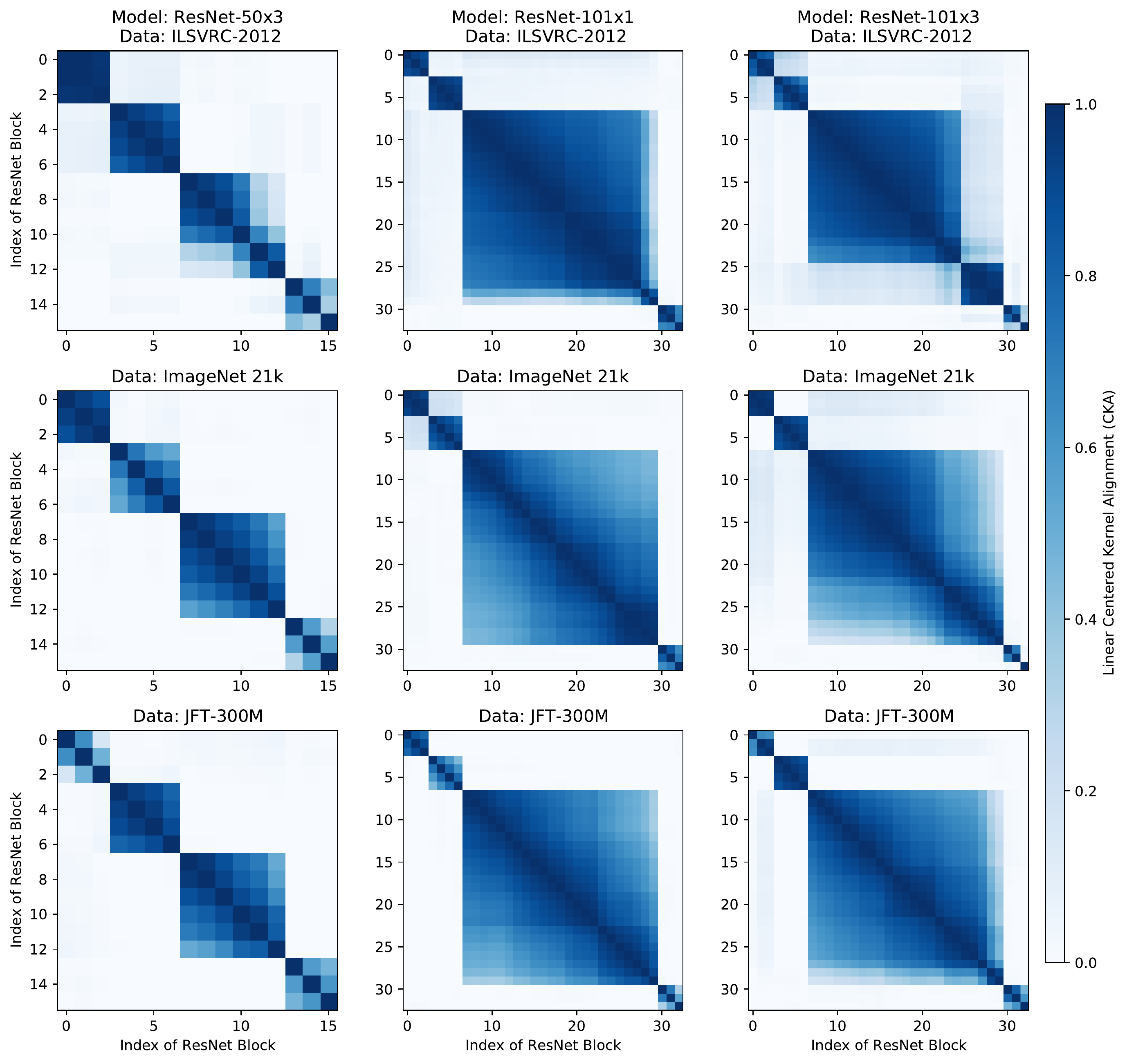}
    \caption{\textbf{Correlation of Representations Across Residual Network Blocks}. We compare the representations (hidden features) after each residual block to those of all other blocks in a residual network. We compare the similarity across representations using Linear Centered Kernel Analysis. This allows for the comparison of representations across stages which are of different shape. We compare three different ResNet models (ResNet-50x3, ResNet-101x1, ResNet-101x3) and three different pre-training datasets (ILSVRC-2012, ImageNet 21k, JFT-300M). The factor `x' represents a multiplier on the number of channels for the ResNet models. All models are fine-tuned on \ione\, and activations are calculated on a random subset of 4096 samples from the ILSVRC-2012 validation set.}
    \label{fig:correlation_study_appendix_resnet}
\end{figure*}

This section includes extended results for layer correlation study. Figure~\ref{fig:correlation_study_appendix_all_tokens} shows the correlation of representations across Transformer blocks for four different ViT models (ViT-B/32, ViT-B/16, ViT-L/32, ViT-L/16) and three different pre-training datasets (ILSVRC-2012, ImageNet 21k, JFT-300M). Figure~\ref{fig:correlation_study_appendix_cls_tokens} shows the correlation for the same models when only taking the CLS token into account. As a comparative reference, Fig.~\ref{fig:correlation_study_appendix_resnet} shows the correlation of representations across residual network blocks for three different ResNet models (ResNet-50x3, ResNet-101x1, ResNet-101x3) and the same three different pre-training datasets (ILSVRC-2012, ImageNet 21k, JFT-300M).

\FloatBarrier  %

\section{Lesion Study}\label{appendix:lesion}
\begin{figure*}[hbt!]
    \centering
    \includegraphics[width=\textwidth]{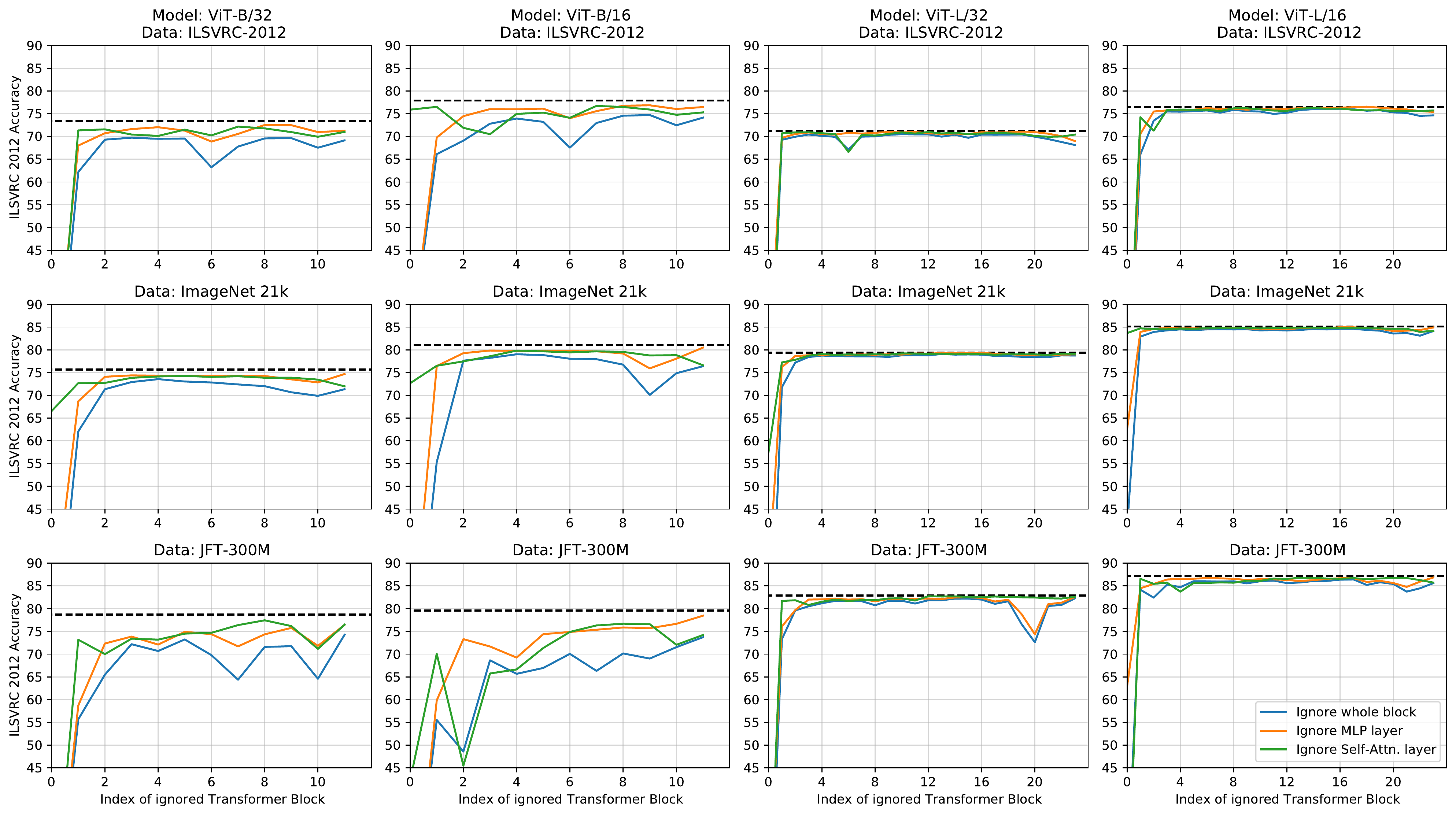}
    \caption{\textbf{Lesion Study}. Evaluation of ViT models when individual blocks, MLP layers or Self-Attention layers are removed from the model after training. We compare four different ViT models (ViT-B/32, ViT-B/16, ViT-L/32, ViT-L/16) and three different pre-training datasets (ILSVRC-2012, ImageNet 21k, JFT-300M). All models are fine-tuned on \ione.}
    \label{fig:lesion_study_appendix}
\end{figure*}

This section includes extended results for lesion study. Figure~\ref{fig:lesion_study_appendix} shows an evaluation of ViT models when individual blocks, MLP layers or Self-Attention layers are removed from the model after training. We compare four different ViT models (ViT-B/32, ViT-B/16, ViT-L/32, ViT-L/16) and three different pre-training datasets (ILSVRC-2012, ImageNet 21k, JFT-300M). All models are fine-tuned on \ione.

\FloatBarrier  %

\end{document}